\documentclass[review]{elsarticle}

\usepackage{lineno,hyperref}
\usepackage[dvipsnames]{xcolor}
\usepackage[margin=1in]{geometry}
\usepackage{subfigure}
\usepackage{tabularx, ragged2e, booktabs}
\usepackage{amssymb}
\usepackage[ruled,vlined,linesnumbered]{algorithm2e}
\usepackage{stmaryrd} % iverson llbracket rrbracket

\SetKwInOut{Input}{Input}
\SetKwInOut{Output}{Output}
\usepackage{amsmath}
\usepackage{mathtools}
\DeclareMathOperator*{\argmax}{arg\,max}
\DeclareMathOperator*{\argmin}{arg\,min}

\newcommand{\ignore}[1]{}

\makeatletter
\def\ps@pprintTitle{%
 \let\@oddhead\@empty
 \let\@evenhead\@empty
 \def\@oddfoot{}%
 \let\@evenfoot\@oddfoot}
\makeatother

\biboptions{authoryear}
\begin{document}

\begin{frontmatter}

\title{Probability estimation and structured output prediction for
learning preferences in last mile delivery}
% \title{Article Title\footnote{The title should be concise, informative, and avoid abbreviations, acronyms, formulae, and cited references to other research works. Please remove this footnote from your actual manuscript.}}

\author[ad1,ad4]{Rocsildes Canoy\corref{cor1}}
\ead{rocsildes.canoy@vub.be}
\author[ad2]{Victor Bucarey}
\ead{victor.bucarey@uoh.cl}
\author[ad4]{Yves Molenbruch}
\ead{yves.molenbruch@vub.be}
\author[ad1]{Maxime Mulamba}
\ead{maxime.mulamba@vub.be}
\author[ad1]{Jayanta Mandi}
\ead{jayanta.mandi@vub.be}
\author[ad1,ad3]{Tias Guns}
\ead{tias.guns@kuleuven.be}
\cortext[cor1]{Corresponding author}

\address[ad1]{Data Analytics Laboratory, Vrije Universiteit Brussel, Belgium}
\address[ad4]{Mobility, Logistics and Automotive Technology Research Centre (MOBI), Vrije Universiteit Brussel, Belgium}
\address[ad2]{Institute of Engineering Sciences, Universidad de O'Higgins, Rancagua, Chile}
\address[ad3]{Dept. Computer Science, KU Leuven, Belgium}

% \author{Gloria Lopez, Li Wan}
% \address{Department of Operations Research, University X}
% \address{410 Terry Ave. North, Seattle, WA, USA}

\begin{abstract}
We study the problem of learning the preferences of drivers and planners in the context of last mile delivery.  Given a data set containing historical decisions and delivery locations, the goal is to capture the implicit preferences of the decision-makers. We consider two ways to use the historical data: one is through a probability estimation method
that learns transition probabilities between stops (or zones). This is a fast and accurate method, recently studied in a VRP setting. Furthermore, we explore the use of machine
learning to infer how to best balance multiple objectives such as distance, probability and penalties.
Specifically, we cast the learning problem as a structured output prediction problem, where training is done by repeatedly calling the TSP solver.
Another important aspect we consider is that for last-mile delivery, every address
is a potential client and hence the data is very sparse. Hence, we propose a two-stage approach
that first learns preferences at the zone level in order to compute a zone routing; after which
a penalty-based TSP computes the stop routing. Results show that the zone transition
probability estimation performs well, and that the structured output prediction learning
can improve the results further. We hence showcase a successful combination of both probability estimation and machine learning, all the while using standard TSP solvers, both during learning and to compute the final solution; this means the methodology is applicable to other, real-life, TSP variants, or proprietary solvers.
\end{abstract}

\end{frontmatter}

\section{Introduction}

The growing trend of end consumers adopting e-commerce and wanting to receive their goods without leaving the convenience of their homes has substantially increased the demand for delivery services. In a logistics and freight transport context, the ``last mile" denotes the final leg of the supply chain where goods are transported and delivered directly to the final recipients. Despite being the shortest phase of the long freight transport chain, it is oftentimes regarded as the most complex and most costly, with last mile deliveries accounting for more than fifty percent of the total supply chain cost. More than ever, it has become pertinent to find more efficient ways to manage last mile delivery systems.

Last mile delivery is complex and unique in its characteristics. One source of complexity is the large number of constantly-changing, geographically-dispersed delivery locations. It is also especially common in practice for route planners to suggest routes that are tailor-made to the specific urban environments, instead of recommending those provided by an off-the-shelf software. The drivers themselves are known to constantly deviate from the planned optimal routes \citep{ceikute2013routing, li2018learning}. Understanding the decision-making behaviors of the drivers and the planners is therefore essential to offer routing plans that capture their preferences and that are `good' according to their subjective evaluation. For that reason, it is crucial to optimize over a measure that represents the preferences of the drivers and planners. Given the difficulty of measuring and defining preferences, a common approach in the literature is to use machine learning to learn these preferences through past decisions, for example in recommender systems~\citep{lu2015recommender} or search engines \citep{joachims2002optimizing}.  

However, our situation is different as the predictions, the output of the method, have to be  TSP solutions.
While neural network approaches have been proposed that can learn combinatorial sequence-to-sequence models \citep{NIPS2015_29921001,kool2018attention},
these methods are known to require huge amounts of data (more than is available in a last mile setting),
and to have difficulty generalizing over arbitrarily sized routings.
Furthermore, we ambition a learning approach that uses a TSP solver as a black box, meaning that
the output is always predicted by a TSP solver. This means that the approach is compatible regardless
of what side constraints are formulated in the TSP solver, and whether it is Mixed Integer Programming-based or a proprietary solver of a company.

On the learning side, we propose a novel combination of two learning approaches: a fast probability estimation method that
estimates the transition probabilities and that can be reformulated such that any TSP solver can be used
to find the maximum-likelihood TSP solution; as well as a structured output prediction method that
can learn the weights of a linear preference function by gradient descent, calling a black-box TSP
solver before every weight update.

A further complication is that in a last-miles setting,  delivery stops are different from one day to another. In other words, there is rarely a repetition between delivery locations on a daily basis. That makes the preference learning task difficult to perform, due to the sparsity of the data. The goal of this article is to provide a methodology to learn the preferences of the planners and drivers when using one route over another taking into account this specific issue. 

\begin{figure}[t]
    \centering
    \includegraphics[scale=0.75]{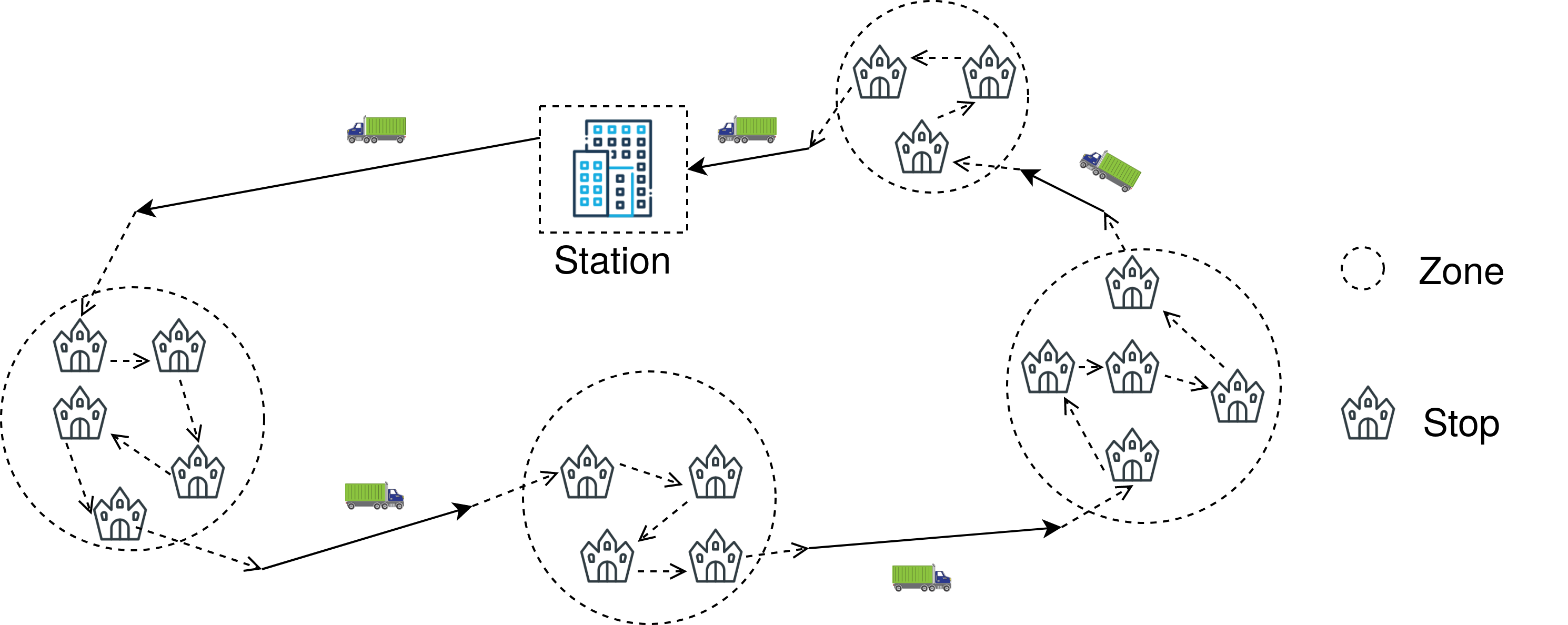}
    \caption{The schematic diagram of the two stage TSP approach with zone and stop ordering}
    \label{fig:schema}
\end{figure}

%In this manuscript, we focus on the challenges in last mile delivery faced by the e-commerce company, Amazon. With access to historical data from multiple delivery stations in North America, our goal is to develop a methodology that can produce good quality delivery routes. Each day, each station receives the delivery locations, for which an efficient route for a single vehicle is to be generated. Solving a TSP for this daily instance is within the reach of existing TSP solvers. The challenge, however, is to extract human knowledge from previous expert-made solutions and use the learned preferences to predict good quality solutions.

%\subsection{Problem definition}

We consider the following setting: there is a set of delivery locations (or \emph{stops}) denoted by $\mathcal{S}$. At our disposal there is a daily historical set of solutions $\mathcal{H} = \{S_t, q^t, x^t\}_{t=1}^T$, where each $S_t \subseteq \mathcal{S}$ is the set of stops at day $t$, and $x^t$ indicates the route plan followed on that specific day. Route plans $x^t$ (or {\emph{solutions}}) are feasible solutions of the travelling salesman problem (TSP) \citep{miller1960integer}. 
% NOT RELEVANT HERE? We consider the binary encoding of a TSP, meaning that each component of $x^t_{ij}$ takes value 1 if the route goes from stop $i \in S_t$ to stop $j \in S_t.$ 
Vector $q^t$ states other relevant features of the data of day $t.$ In particular, we consider an optional special attribute namely the drivers' subjective {\it route quality score} which takes values in  $\{\mathtt{high,\, medium,\, low}\}$. %, which indicates a user evaluation of the route.

The majority of the stops appear only a few times in an $S_t$, making it non-viable to directly learn relations between the stops in $\mathcal{S}$. However, stops are partitioned into zones $\mathcal{Z} = \{Z_1, Z_2, \ldots, Z_m\}$ of the whole territory. This zonification is represented as a partition of $\mathcal{S}$. In other words, $\bigcup_{k=1}^m Z_k = \mathcal{S}$ and $Z_k \cap Z_{k'} = \emptyset$ for $k\neq k'$. This situation is depicted in Figure \ref{fig:schema}. 
As zones cluster stops together, the historical data will have frequently re-occuring zone-to-zone transitions creating the possibility to learn zone transition preferences.%, have much more cases where  in which it is assumed that the set of historical solutions at zone level contains more transitions of the form {\it go from zone $Z_k$ to $Z_k'$} than transitions at the stops level. 
The goal is to build a model such that given a new set of stops $S$ the model returns a routing plan $x$ that largely follows the zone-level preferences of the planners and drivers as well as taking the total route distance into account. 

We propose a two-stage TSP approach to solve this problem: one TSP over the zones, and a second one over the stops that takes into consideration the sequences of zones. These two TSPs involve three learning processes: two to learn preferences over zones and one to learn good zone \textit{penalties} when solving the stop-level TSP.

\begin{figure}[t]
    \centering
    \includegraphics[scale=0.45]{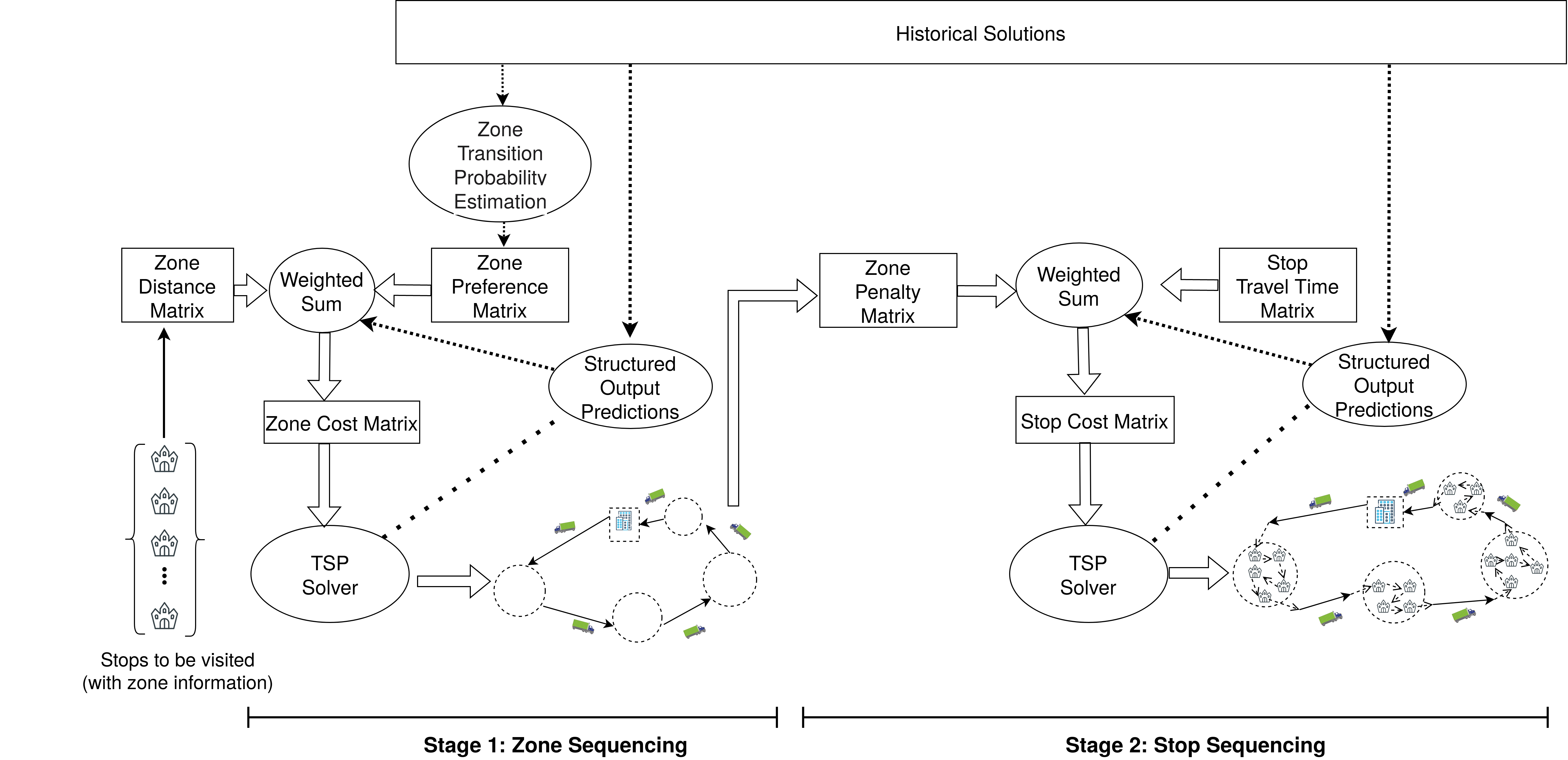}
    \caption{Diagram of the proposed methodology}
    \label{fig:approach}
\end{figure}

A diagram of the methodology is depicted in Figure \ref{fig:approach}.
%To do so, our methodology is summarized as follows: 
We first assume that preferences can be captured as \emph{costs} of going from one zone/stop to another zone/stop. This assumption is standard as it is used in \citet{canoy2019CP, mandi2021data, chen2021inverse}. One of the advantages of this assumption is the possibility to use exact and/or approximate algorithms available for the TSP. We then estimate zone transition probabilities from historical data, as well as learning the best weighted combination of the zone distance matrix and the zone preference matrix, using \emph{structured output prediction} \citep{sop_book}. 
With this zone cost matrix, a sequencing of zones is obtained by solving a zone-level TSP. % between the zones using the learned vector cost $\Tilde{c}_{ij}$. 
In the next stage, we use the computed zone sequence to define a zone penalty matrix, penalizing stop-level transitions that do not respect the computed zone ordering. The weights of these penalties are again learned using structured output prediction. Together, this forms the stop cost matrix, where a TSP solver is used to computed the routing over all stops (potentially with time-windows and other operational constraints).

Thus, in our approach there are two TSPs to solve: the first one to determine a sequence of zones, and the second one to route over the stops considering the zone sequencing. 

%Delivery locations (or \emph{stops}) are uniquely labeled, . The majority of the stops, however, are assigned into zones, which provide an intrinsic clustering of related stops. Hence, there is an opportunity to learn preferences between zones.

% Key technical challenges:
% \begin{itemize}
%     \item The TSP instances are within the reach of existing TSP solvers, the key issue is to learn/predict which sequencing will be more preferred
%     \item Most stops are unique (can you provide some statistics?), leaving little room to learn relations between individual stops
% \end{itemize}

% However, most stops belong to a zone, hence zones provide an intrinsic clustering of related stops. Furthermore, for different training instances, even when from different depots, stops from the same zones are visited. Hence there is opportunity to learn preferences between the zones.

%More specifically, we will assume that the underlying TSP solution method is a zone-based TSP approach, that is, a method that penalizes traveling from one zone to another. The key challenge is then to learn the penalties for moving between the different zones. For this, we will build on our earlier work \citep{canoy2019CP} where we replace the distance matrix between stops by a learned probability matrix. More specifically, we take a first order Markov approximation of a routing probability, and use a counting-based estimation approach that works well even in limited data settings.

The contributions of our work are the following:
\begin{itemize}
    \item We propose a two-stage TSP approach where we learn probabilistic preferences between zones, and use these preferences in defining the penalties in the TSP.
    \item We estimate transition probabilities with a Markovian counting method, and augment it with structured output prediction \citep{sop_book} to learn appropriate penalties that trade-off distances and learned preferences at the zone level.
    \item We compute a zone-ordering but do not enforce it as hard constraints. Instead, we penalize zone-order violations and \textit{learn} the penalties over historic training data.
    \item We experimentally show the suitability of learning only at the zone level, as well as the benefits of structured output prediction to learn both at the zone level and stop level, on the real data set.
\end{itemize}

% \textbf{T: something like this?}
% \textbf{The biggest 'open' part I think is different ways to integrate the penalties in the TSP, even e.g. directly; and in extension maybe even doing the SOP on that direct formulation (if runtime would allow that?)}
% \textbf{For me, an elephant in the room here is: why don't we use our neural network approach instead of SOP? or even a such a simple neural network inside the SOP?}

\section{Literature Review}

The \emph{last mile delivery}~\citep{zeng2019last} problem has been well studied in the literature. Because of its extensive use in e-commerce, it has received increasing attention in recent years. Last mile delivery may be viewed as a travelling salesman problem (TSP)~\citep{miller1960integer} or a vehicle routing problem (VRP) \citep{dantzig1959truck}, for which there are many existing available solvers.
% As it involves multiple stakeholders, each having a different objective function, different routings are possible depending on which has been considered as the objective. For instance, a common approach is to minimize the time to finish serving all customers, i.e., the makespan of all the routes \citep{yu2017minimum}. On the other hand, \citet{das2018minimizing} approach the problem with the objective of minimizing latency or the waiting time of the customers.
  
However, \citet{li2018learning} state that in most cases, the last mile delivery does not go according to the planned schedule. In the context of routing between point to point, \citet{ceikute2013routing} highlight that drivers tend to follow frequently used routes, which are not necessarily shorter in terms of distance or travel time. This is primarily because, in deciding their routes, drivers consider factors which are not in the objective function, e.g., traffic congestion and availability of parking and fuel stations. Drivers have tacit knowledge of this information. \citet{toledo2013decision} also point out that drivers tend to rely on tacit knowledge to plan their routes. Explicit formalization of their knowledge poses an insurmountable task.

Several works have studied the problem of learning the routing preferences of the drivers from past routing trajectories. \citet{delling2015navigation} present a customizable route planning framework that infers the preferences of the drivers from GPS traces by learning individualized cost functions. \citet{letchner2006trip} compute the ratios of the individual drivers' travel time and the theoretical optimal travel time to learn the implicit preferences and biases. 
Both of the above-mentioned studies focus on point to point commute for individual drivers. %In our work, we do not explicitly learn the parameters of the cost function nor do we estimate preferences from the travel times. 

\citet{canoy2019CP} approach the problem in a vehicle routing setting from a different perspective---they introduce a weighted Markov model to learn the preferences from historical routes. This technique avoids the need to explicitly specify the preference constraints and implicit sub-objectives. Their Markovian model computes the transition probabilities between the stops. The model is then used to solve a \emph{maximum likelihood routing} problem to come up with a routing scheme. We extend this framework to develop a two-stage TSP, where in the first stage we adapt the technique to compute the transition probabilities between zones to find the zone order. %In the following section, we provide a formalisation of the TSP, as it is crucial in the presentation of our proposed algorithms.

% \begin{itemize}
%     \item TSP - \textit{Integer Programming Formulation of Traveling Salesman Problems,} \cite{miller1960integer}
%     \item VRP - \textit{The Truck Dispatching Problem,} \cite{dantzig1959truck}
%     \item Last-mile delivery - \textit{Last-Mile Delivery Made Practical: An Efficient Route Planning Framework with Theoretical Guarantees,} \cite{zeng2019last}
%     \item Driver behavior vs recommended route - \textit{Routing Service Quality -- Local Driver Behavior Versus Routing Services,} \cite{ceikute2013routing}; \textit{Learning from Route Plan Deviation in Last-Mile Delivery,} \cite{li2018learning}
%     \item Learning from historical solutions - \textit{Vehicle Routing by Learning from Historical Solutions,} \cite{canoy2019CP}
% \end{itemize}

\section{Background: the Traveling Salesman Problem (TSP)}  Given a set $S$ of stops, with (assymetric) \emph{costs} $c_{ij}$ associated to each $(i,j)$-pair of stops, the TSP consists of determining a minimum cost circuit passing through each vertex once and only once \citep{miller1960integer, laporte1992traveling}. An integer programming formulation to solve the TSP is:
\begin{align}
    \mathrm{(TSP)}\quad \,& \min \sum_{i\neq j} c_{ij}x_{ij} \label{eqn:obj_cost}\\
    \text{subject to} \quad & \sum_{j\in S} x_{ij} = 1 & i\in S,\ i\neq j \label{eq:flow1} \\
    & \sum_{i\in S} x_{ij} = 1 & j \in S,\ j\neq i  \label{eq:flow2} \\
    &\sum_{i,j \in V} x_{ij} \le |V|-1 & V\subset \{1,\ldots, n\},\ 2\le |V| \le  |V|-2  \label{eq:subtour} \\
   & x_{ij} \in \{0,1\}& i,j \in S,\ i\neq j  \label{eq:natx}
\end{align}

\noindent where the variable vector $x$ is defined as above. The objective function minimizes the sum of the cost over the arcs in the route. The first two constraints ensure that each stop is visited onle once. The third constraint ensures that there are no subtours in the solution. 

Our approach does not rely on an integer programming formulation, but works with any TSP solver. By abuse of notation, we will refer to TSP$([c_{ij}])$ as an anytime oracle that returns an optimal solution of (TSP) when the cost vector $c$ is optimized, or the best one found in case a timelimit is reached.

Several exact methods and heuristics are being developed to solve this problem since the 60s, making it one of the most studied problems in operations and transportation research \cite{applegate2011traveling}. 

\section{Methodology}

%In this section we describe in detail how we perform the two-stage TSP. The first learning stage, learning the preferences of transitions between zones, consists in mixing the distances between zones and a function of how many times they appear in the historical set of solutions  $\{S_t,  q^t, x^t\}_{t=1}^T$. This second term can be interpreted as a transition probability between zones as it is referred in \cite{canoy2019CP}. The second stage consists in learning the penalties of not respecting the order found at stage 1. 

In this section we describe in detail our methodology to learn and implement the two-stage TSP. Figure~\ref{fig:approach} shows the overview, where the first learning stage, {\it zone ordering}, consists in catching preferences over transitions between zones. These preferences are modelled as a linear combination of two terms: the first one related to the distance between zones, and a second one which is a function of the frequency with which each transition appears in the data training. We then apply an {\it structured output prediction} approach to learn the weights in the linear combination. The second stage, {\it stop ordering}, consists in learning how to route over stops taking into account a predefined zone ordering. This task is performed by setting penalties each time that a solution does not respect the zone ordering. The values of this penalties are also learnt by structured output prediction.    

% \begin{scriptsize}
% \begin{itemize}
%     \item Introduce two-stage algorithm. Overview of the goal(s) of each stage.
%     \item Stage 1: Zone ordering
%     \begin{itemize}
%         \item Distance-based ordering
%             \begin{itemize}
%                 % \item Treatment of stops with ``no zone"
%                 \item TSP on absolute distances
%                 \item Zone location approximation (centroid)
%                 \item TSP on normalized distances
%             \end{itemize}
%         \item Learning zone order from historical data
%             \begin{itemize}
%                 \item Introduce Markov model from literature
%                 \item Numerical equivalent of each route score (\texttt{High, Medium, Low})
%                 \item Transition probability matrix construction algorithm
%                 \item TSP on transition probabilities
%             \end{itemize}
%         \item Distance + probabilities mixing
%             \begin{itemize}
%                 \item Introduce structured output prediction (SOP)
%                 \item Learning optimal convex combination parameters by SOP
%                 \item TSP using learned parameters to order zones
%             \end{itemize}
%     \end{itemize}
%     \item Stage 2: Stops ordering
%     \begin{itemize}
%         \item TSP using given travel time matrix + zone penalty
%         \item Setting zone penalty parameters
%         \item Combining travel time + zone transition probabilities from Stage 1 (via SOP?)
%     \end{itemize}
% \end{itemize}
% \end{scriptsize}

\subsection{Stage 1. Zone ordering.}

The first phase of our proposed two-stage approach is the zone ordering phase. Given the set of stops (delivery locations), with each stop assigned to a zone, our goal at this stage is to generate a good zone order.
Note that zones need not be supplied by the user per se, one could use a clustering method, or geographical regions too. The main point is to make the number of regions of interest (stops/zones) less sparse, so that learning a function that can be applied in new unseen situations becomes more meaningful.

\subsubsection{Zone order by distance} A direct approach to get a sequence of zones is to use the distance between them, and minimizing the distance of visiting all of them. Given a set of stops $S$ and the corresponding set of zones $Z_1, \ldots, Z_{m}$ that contain the stops, we solve a TSP to visit each zone. In this case, we use as a cost $c_{ij}$ in Equation \eqref{eqn:obj_cost} the euclidean distance between the centroid of zone $Z_i$ and $Z_j$, noted by $d_{ij}$. Formally, if the stops inside the zone $Z_i = \{s_1, \ldots, s_{|Z_i|}\}$ are geolocated with coordinates $q^{long}(s), q^{lat}(s)$, then the centroid is given by:

\begin{equation}
    q^{long}(Z_i) = \frac{1}{|Z_i|} \sum_{s \in Z_i} q^{long}(s),   \quad    q^{lat}(Z_i) = \frac{1}{|Z_i|} \sum_{s \in Z_i} q^{lat}(s).
\end{equation}

\noindent As the coordinates of the stops within a zone are relatively close to one another, we treat the earth as being locally flat and determine the centroid by taking the average of the latitudes and the longitudes of all the stops. Then, the distances $D = [d_{ij}]$ are given by

\begin{equation}
    d_{ij} = \sqrt{\left(q^{long}(Z_i)-q^{long}(Z_j)\right)^2 + 
                    \left(q^{lat}(Z_i)-q^{lat}(Z_j)\right)^2 }
\end{equation}

Note that this zone sequencing by distance approach does not make use of the historical data. There is no learning involved, hence the method is able to provide a solution for any given instance without any model training. While this traditional approach indeed gives the distance-wise optimal zone sequence, there is no guarantee that the resulting order is close to that of the expert-made solution.

\subsubsection{Zone order using transition probabilities}

\paragraph{\bf Learning zone order from historical data}  We adapt the approach introduced in \cite{canoy2019CP} of learning transition probabilities between zones from the historical data. This approach models preferences as a Markov chain, where the preferences are learnt as the probability of going from one zone/stop to another. The idea is that by determining the zone transition probabilities, we are implicitly learning the latent preferences of the expert. 

Let us denote $\mathbf{P}$ the transition probability matrix, where each component $P_{ij}$ represents the probability of going from zone $Z_i$ to $Z_j$. While the probabilistic model is relatively simple, it allows us to construct a zone transition probability matrix. Solving (TSP) using this transition matrix will give us the \emph{maximum likelihood zone sequence}. This maximum likelihood problem can be cast as:

\begin{equation}
    \max_{x} \prod_{(i,j)} P_{ij} x_{ij}: \quad x \mbox{ satisfies \eqref{eq:flow1} - \eqref{eq:natx}} ,
\end{equation}
 
\noindent which is equivalent to solve a TSP with $c_{ij} = -log(P_{ij})$ in \eqref{eqn:obj_cost}, as $\max_{x} \prod_{(i,j)} P_{ij} x_{ij} = \min_{x} \sum_{(i,j)} -log(P_{ij}) x_{ij}$. So the model is compatible with any TSP solver. Now we focus on how the transition probability matrix $P$ is obtained.

%In order to generate zone sequences that are close to those of the solutions made by the expert, we adapt the approach introduced in \cite{canoy2019CP} of learning transition probabilities between zones from the historical data. The idea is that by determining the zone transition probabilities, we are implicitly learning the latent preferences of the expert.

\begin{algorithm}[htbp]
% \scriptsize
\SetAlgoLined
\Input{Set $\mathcal{H}= \{S_t, q^t,  x^t\}_{t=1}^T$ of actual traversed routes from historical data; %each route $x^t$ labeled with route quality score $q^t\in \{\mathtt{high, medium, low}\}$\\
Optionally, per-instance weights $v^t,$ that influence the importance of the instance.%which are numerical equivalents of the labels $q$
}
\Output{Zone transition probability matrix $\mathbf{P}$ }
% Extract and gather all the zones visited in $R_S$ into a set $\Sigma = \{z_0,z_1,\ldots,z_t\},$ where $z_0$ denotes the zone of station $S$.\\
 \For{\emph{each} $x^t,$ $t=1,\ldots,T$}{
    %Determine the sequence of zones $Z_{r_k}$ traversed in $r_k.$\\
    Construct an adjacency matrix $\mathbf{A}^t = [a^t_{ij}],$ where $a^t_{ij}=1$ if the transition $(Z_i\rightarrow Z_j)$ is in $x^t,$ and $0$ otherwise.\label{line_form_adjacency}}
Build the zone transition \textit{frequency} matrix $\overline{\mathbf{P}} = [\overline{p}_{ij}]$ by (weighted) counting over the the adjacency matrices constructed in the previous step
\begin{equation}
   \overline{\mathbf{P}} = \sum_{k=1}^p v^t\mathbf{A}^t.
\end{equation}\\
Normalize rows of $ \overline{\mathbf{P}}$ by dividing each element by the row sum to obtain the zone transition probability matrix:
\begin{align*}
           p_{ij} &= \frac{\overline{p}_{ij}}{\sum_{k} \overline{p}_{ik}}.
\end{align*}

 \textbf{return} Zone transition probability matrix $\mathbf{P} = [p_{ij}],$ where $
    p_{ij} = \text{probability of } (Z_i\rightarrow Z_j)$
 \caption{Building the zone transition probability matrix %for a \emph{single} station
 }\label{ZoneTransAlgo}
\end{algorithm}

\paragraph{\bf Transition probability matrix construction algorithm} Algorithm~\ref{ZoneTransAlgo} outlines the process of constructing the zone transition probability matrix from historical data. Let $\mathcal{H} = \{S_t,  q^t, x^t\}_{t=1}^T$ denote the set of historical routes and $\mathcal{Z} = \{Z_1, \ldots, Z_m\}$ the zonification. 

We build a transition probability matrix $\mathbf{P} \in [0,1]^{m \times m}$ by computing a {\it weighted } frequency of appearance of each transition between zones. The algorithm works as follow: For each route $x^t$, an adjacency matrix $\mathbf{A}^t$ is built where $a^t_{ij}$ equals to 1 if the transition $Z_i\rightarrow Z_j$ occurs, and 0 otherwise. Then, a matrix $\overline{\mathbf{P}}$ is built by summing over all the adjacency matrices weighted by instance-specific parameters $v^t$. Finally, the transition probability matrix $\mathbf{P}$ is obtained after normalization. 

When a route quality score, e.g., $\{\mathtt{high,\, medium,\, low}\},$ is associated to each route, we can control how much the zone sequence contained in each route influences the resulting zone transition probability matrix (step 4 of Algorithm \ref{ZoneTransAlgo}). This can be achieved by allocating a larger weight to $\mathtt{high}$ quality routes (and a smaller weight to $\mathtt{low}$ quality routes). For example, we can assign numerical weights $\{v_{\mathtt{high}},\, v_{\mathtt{medium}},\, v_{\mathtt{low}}\}$ corresponding to each of the score labels $\{\mathtt{high,\, medium,\, low}\},$ where $v_{\mathtt{high}}\ge v_{\mathtt{medium}}\ge v_{\mathtt{low}}.$ Alternatively, we can train selectively with only the $\mathtt{high}$ quality routes by setting $v_{\mathtt{high}}=1$ and $v_{\mathtt{medium}} = v_{\mathtt{low}}=0,$ etc.

\subsubsection{Zone order by mixing distances and transition probabilities}

We take inspiration from \cite{canoy2019CP} where experimental results show that a cost matrix that mixes (1) distances and (2) transition probabilities, can lead to better solutions being found. From an application point of view, this is motivated by the observation that planners have personal and experience-based preferences, but also always take the total distance into account. From a probabilistic inference perspective, it could be that the estimated probability matrix is uncertain (or indifferent) with respect to certain stops/zones, for example, new stops that have rarely (or even not) been visited before. This too suggests it can be beneficial to combine distances with the estimated probabilities.

The formulation of the cost matrix $[c_{ij}]$ for the TSP is then:
\begin{align}
	c_{ij} = -w_d log(d'_{ij})  -w_p log(p_{ij}) \label{eq:mix_cost}
\end{align}
where $\mathbf{P}$ is the estimated preference matrix and $\mathbf{D}'=[d'_{ij}]$ is a distance-based probability matrix. %For both, each of the row-sums sum up to 1 and hence they can be meaningfully combined.

In our distance-based probability, we want stops that are closer together (so with small $d_{ij}$) to have higher probability than stops that are far apart. We hence first invert the distance and then normalize the rows so they sum up to $1$:
\begin{align}
    d'_{ij} = \frac{ 1/d_{ij} }{ \sum_k 1/d_{ik} }.
\end{align}

Both values are in the interval $[0,1]$, but it is nonetheless impossible to know upfront how to best mix them, that is, what values of $(w_d, w_p)$ will result in a cost matrix $[c_{ij}]$ leading to the best zone routing.

We propose a principled way to learn the $(w_d, w_p)$ values from historic data. The key challenge here is that we do not know what the \textit{true} cost matrix $[c^*_{ij}]$ is. If this would have been the case, we could have treated the weight learning as a standard machine learning regression problem.

Instead, we only know for each historic routing, what the followed zone tour $x^*$ was; and that a given $(w_d, w_p)$ tuple leads to a cost function $[c_{ij}]$ for which we can compute the optimal zone tour using a TSP solver. In other words, the actual values of the predicted $[c_{ij}]$ do not matter, only the resulting tour x as the optimal solution of TSP$([c_{ij}])$. We observe that this can be seen as a \textbf{structured output prediction problem}.

Structured output prediction is used in a setting where given an input, we want to predict an output that satisfies a predefined \textit{structure}. For instance, in case of part of speech (PoS) tagging,  the input is a sentence and the output is the sequence of PoS tags, which must have the combinatorial and compositional structure of the English language. In this case, treating the PoS tag of each word independently  ould be inefficient as such an approach would fail to capture the expressivity of the language. The structured output prediction approach, on the other hand, predicts PoS tags of multiple words simultaneously and utilizes the dependencies between them to maintain the \emph{linguistic structure} in the output~\cite{sop_book}.
% A structured output prediction problem~\cite{sop_book} is a machine learning problem where the output to predict satisfies a predefined \textit{structured}, for example that it forms a sequence, tree or graph. 
Structured output prediction has been successfully used in complex applications such as semantic parsing in natural language processing~\citep{ClarkeGCR10}, semantic role labeling~\citep{YadollahpourBS13} and computational biology~\citep{Chen0UGS20}.

In our case, the predefined structure is that the prediction has to be a circuit that visits every stop exactly once: a feasible TSP solution. In general, in structured output prediction, the problem of searching for a score-maximizing output structure (called the \textit{inference} task) is the following:
\begin{align}
	\hat{x} = \argmax_{x \in \mathcal{X}(u)} f(x; \mathbf{w}, u)
	%F(x) = \argmax_{y\in\mathcal{Y}(x)}\mathbf{w}^T\Phi(x,y),
\end{align}
where $u$ is the input (in our case: the set of zones to route over, as well as the pairwise distance and any other relevant information), $\mathcal{X}(u)$ is the (implicit) space of all possible outputs that satisfy the structural constraints (e.g. the feasible solutions of the TSP defined implicitly by \eqref{eq:flow1} - \eqref{eq:natx}) and $\mathbf{w}$ are the model parameters that we wish to learn. Without loss of generality we will assume that $f$ needs to be maximized to stay close to the traditional structured output prediction literature, (In contrast to the literature, we write $(u,x)$ to denote input/output structure instead of $(x,y)$ as $x$ is more natural for the TSP (output) variables in this paper), even though the TSP is formulated to minimize its cost matrix (or equivalently maximize its -cost matrix).

Many approaches to structured output prediction use approximate inference techniques, for example using the Viterbi algorithm to decode a sequence~\citep{sop_book}. However, TSPs have a peculiar structure and very efficient specialised solvers that are often \textit{anytime}, meaning they can be interrupted with a timeout and will return the best solution found so far. We hence do not mind calling a TSP solver repeatedly during learning.

We will use the seminal structured perceptron algorithm of Collins~(see \cite{collins2002discriminative}). It is a variant of the perceptron algorithm, which in turn forms the basis of neural network learning. While it is restricted to learning a linear function over its input, this is sufficient for our needs. It furthermore has nice convergence properties in case of separable data, similar to the perceptron algorithm~\citep{collins2002discriminative}. 

%The structured perceptron algorithm learns a mapping from inputs $x\in\mathcal{X}$ to outputs $y\in\mathcal{Y}.$ We assume:

We assume given a dataset $\mathcal{H} = \{(u,x)\}$ of $T$ training examples with $u$ corresponding to an input structure, and $x$ the intended output structure, where $x \in \mathcal{X}(u)$.

The structured perceptron algorithm requires to define a problem specific \textbf{representation} function $\Phi$ that maps every valid $(u,x)$ to a feature vector $\Phi(u,x)$. The length of the feature vector determines the length of the weight vector $\mathbf{w}$, in such a way that the linear function $\mathbf{w}^\intercal\Phi(u,x)$, where $\mathbf{w}^\intercal\Phi(u,x)$ is the inner product $\sum_s w_s\Phi_s (u,x)$ for a feature vector of length $s$, represents the quality of the input-output pair $(u,x)$. In other words, the structured perceptron algorithm assumes that $f(x; \mathbf{w}, u) = \mathbf{w}^\intercal\Phi(u,x)$ such that the inference task becomes:
\begin{align}
	\hat{x} = \argmax_{x\in\mathcal{X}(u)} \mathbf{w}^\intercal\Phi(u,x)
\end{align}

The learning task is to set the weight values $\mathbf{w}$ using the training examples as evidence.

Before looking at the structured perceptron algorithm, we will translate its required input to our TSP setting:
\begin{itemize}
	\item $x$ is a TSP solution, represented by an $n \times n$ adjacency matrix $x=[x_{ij}]$;
	\item $u$ is the input structure, namely the tuple $(n,\mathbf{D}', \mathbf{P})$ with $n$ is the number of stops in the instance, $\mathbf{D}'$ the normalized distance matrix between the $n$ stops, and $\mathbf{P}$ the estimated probability matrix between the $n$ stops;
	\item $\mathbf{w}$ is the length-2 weight vector $\mathbf{w} = [w_d, w_p]$ that trades-off the importance of the distance and the probability in determining the TSP quality;
	\item $\Phi(u,x)$ hence needs to compute a length-2 feature vector such that $\max \mathbf{w}^\intercal\Phi(u,x)$ corresponds to $\min \sum_{ij} -w_d log(d'_{ij})  -w_p log(p_{ij})$ from Equation~\eqref{eq:mix_cost}. We define
\begin{align}
  \Phi\left(u,x\right) 
  &= \Phi\left((n, [d'_{ij}], [p_{ij}]),[x_{ij}]\right) \\ 
  &= \left[\sum_{ij} log(d'_{ij})x_{ij}, \sum_{ij} log(p_{ij})x_{ij}\right]
\end{align}
\end{itemize}

We can now show that
\begin{align}
	\hat{x} &= \argmax_{x\in\mathcal{X}(u)} \mathbf{w}^\intercal\Phi(u,x) \\
	        &= \argmax_{x\in\mathcal{X}(u)} w_d\sum_{ij} log(d'_{ij})x_{ij} + w_p\sum_{ij} log(p_{ij})x_{ij} \\
	        %&= \argmax_{y\in\mathcal{Y}(x)} \sum_{ij} (w_d log(D'_{ij}) + w_p log(\mathbf{P}_{ij}))y_{ij} \\
	        &= \argmin_{x\in\mathcal{X}(u)} \sum_{ij} (-w_d log(d'_{ij}) - w_p log(p_{ij}))x_{ij} \\
	        &= \mbox{TSP}([c_{ij}]) \quad \mbox{with} \quad c_{ij} = -w_d log(d'_{ij})  -w_p log(p_{ij})
\end{align}

The structured perceptron algorithm learns $\mathbf{w}$ given a dataset $\mathcal{H} = \{(u,x)\}$ and a representation function $\Phi(u,x)$. The pseudo-code is shown in Algorithm~\ref{alg:PPVRP}.

\begin{algorithm}[htbp]
	% \scriptsize
	\SetAlgoLined
	initialize $\mathbf{w}$\\
	\For{$e = 1,\ldots,E$}{
		\For{\textbf{\emph{all}} $(u,x)\in \mathcal{H}$}{
			$\hat{x}\leftarrow \argmax_{\hat{x}\in\mathcal{X}(u)}\mathbf{w}^\intercal\Phi(u,\hat{x})$\\
			\If{$\hat{x}\neq x$}{
				$\mathbf{w} \leftarrow \mathbf{w}\, +\, 
				\delta\left(\Phi(u,x) - \Phi(u,\hat{x})\right) $
			}
		}
	}
	\Return{$\mathbf{w}$}
	\caption{The Structured Prediction Algorithm}\label{alg:PPVRP}
\end{algorithm}

The weight vector is initialized on line 1, e.g. to constant values such as $0$ or $1$.   %\tias{WHAT IS H? the number of epochs I would assume, but not explained in the original text?}
In every epoch $e$, the algorithm then iterates over the training examples in $\mathcal{H}$ (line 3). It first computes the predicted TSP $\hat{x}$ of the given instance for weight vector $\mathbf{w}$. If the predicted output does not match the intended TSP $x$, then a small perceptron update to the weights $\mathbf{w}$ is performed, using the difference between the representation function $\Phi(u,x)$ of $u$ with the true structured output $x$ and $\Phi(u,\hat{x})$ of $u$ with the predicted output $\hat{x}$. The $\delta$ is the \textit{learning rate}: a small value that controls how large or small the weight updates are in each iteration. This procedure is repeated for $E$ epochs.

In the experiments, we will evaluate in how far the use of structured output prediction on the weights $\mathbf{w}$ can improve the zone ordering sequence found.

\subsection{Stage 2. Stop ordering.}
Up to now we discussed how in stage 1 we can \textit{estimate} the transition probability between zones from historical data, and how we can \textit{learn} the trade-off between distance and learned probability to obtain good zone-level TSP tours. The goal in this second stage is to find a TSP tour over all the stops, taking the predicted zone ordering of the previous stage into account.

%The goal in this phase is to find an efficient order of stops while respecting the zone order computed in the previous stage.

\subsubsection{Stop order by minimizing travel time}

\noindent
A direct approach in finding a candidate stop order for a given instance, i.e., a set of stops, is by solving (TSP) with the stops as the vertices, and the given distance or  travel time matrix as the cost matrix. Again, here, only the pairwise travel times are taken into account; there is no \emph{clustering,} as the zones associated to the stops are not considered, nor is there any learning from historical data. Hence, while this method results in a stop sequence that is optimal in terms of the total travel time, there is no assurance that its results are close to the ones made by the expert.

\subsubsection{Stop order with zone penalties}

At this stage, we propose to add to the travel times between stops some \emph{zone penalty}, which we can decompose into different penalties for different degrees of order violation compared to the zone ordering of the previous stage.
%define depending on how strictly we want to impose the zone order obtained in the previous stage. 
% For instance, we may want the final stop sequence to strictly follow the zone order from stage 1, e.g., the zone order is regarded as a hard constraint during stop order optimization.

% \medskip
% \noindent
% Let ZO be the zone order obtained in stage 1. For a given pair of stops $(i,j),$ let  $\texttt{zone\_idx\_dist}_{ij} = \texttt{idx(zone$_j$)} - \texttt{idx(zone$_i$)},$ where $\texttt{idx(zone$_k$)}$ denotes the index in ZO of the zone where stop $k$ belongs. Then we have

% \begin{equation*}
%     \texttt{zone\_penalty}_{ij} = 
%     \begin{cases}
%       0, & \text{if $\texttt{zone\_idx\_dist}_{ij} = 0$} \\
%       1, & \text{if $\texttt{zone\_idx\_dist}_{ij} = 1$} \\
%       2, & \text{otherwise.}
%     \end{cases}
% \end{equation*}

%For this implementation, l
Let us suppose the following:
\begin{itemize}
    \item ZO, the zone order obtained in stage 1. This function returns ZO$(Z) = k$ if the zone $Z$ is visited in the k-th order. 
    \item $S=\{s_0,s_1,s_2,\ldots,s_n\},$ the set of stops that need to be routed, with $s_0$ denoting the depot
    %\item $\mathtt{idx(Z_{s_k})},$ the index in ZO of the zone where stop $k$ belongs
    \item $\mathbf{D} = [d_{ij}],$ a travel time matrix whose $(i,j)$-th element represents the time to travel from stop $s_i$ to stop $s_j$ 
\end{itemize}

To find our desired stop order, we minimize the total travel time while giving some penalty for each transition from one zone to the next. The \emph{weight} of these penalties will depend on the zone order ZO obtained in stage 1.

We propose to formulate the cost matrix $[c_{ij}]$ for the second stage TSP as a weighted combination of the distance $d_{ij}$ between stop $i$ and $j$ and how the zones of the two stops are related to the computed zone ordering.

\newcommand{\zn}[1]{\mbox{\texttt{#1}}}

The zone ordering induces an order, also on the stops. Let us define the order index function $O:S \rightarrow \{1,\ldots, m\}$, where $O(s_i)$ (or simply $O_i$) is equal to $k$ if $s_i\in Z$ and ZO$(Z)=k$. Thus, $O_i$ indicates the order where a stop has to be visited. We can now check that two stops $i,j$ belong to the same zone, by checking that $O_i == O_j,$ which is equivalent to checking $Z_i == Z_j$. Furthermore, we can check whether the zone of stop $j$ is the `next' zone in the zone order, compared to the zone of $i$, in which case: $Z_j == Z_i + 1$ is true. Similarly, we can compute whether $j$ is in the `previous' zone according to $ZO$, or two zones ahead, etc.

%Let $Z_i$ be the zone of stop $i$, then $O_i = idx(Z_i, ZO)$, that is, where in the order of zone order $ZO$ that this zone is located. For example, let $ZO = [\zn{r}, \zn{a}, \zn{t}, \zn{g}]$ and let the depot be in zone $\zn{r}$, then $O_depot = 0$. For a stop $i$ with $Z_i = \zn{t}$ we have $O_i = 2$. 

Using the zone order index $O_i$ and the logical expressions just introduced, we can now define a penalized cost matrix for the stop-level TSP as follows:
\ignore{
\begin{align}
	c_{ij} = &\mathbf{w}_0 d_{ij} + \mathbf{w}_1 \llbracket O_i == O_j \rrbracket, \nonumber \\
	         &\mathbf{w}_2 \llbracket O_j == O_i + 1 \rrbracket +
	          \mathbf{w}_3 \llbracket O_j == O_i + 2 \rrbracket +
	          \mathbf{w}_4 \llbracket O_j \geq O_i + 3 \rrbracket \nonumber \\
	         &\mathbf{w}_5 \llbracket O_j == O_i - 1 \rrbracket +
	          \mathbf{w}_6 \llbracket O_j == O_i - 2 \rrbracket +
	          \mathbf{w}_7 \llbracket O_j \leq O_i - 3 \rrbracket 
	\label{penalty_obj}
\end{align}}

\begin{align}
	c_{ij} & =  \mathbf{w}_0 d_{ij} + \mathbf{w}_1 \llbracket O_i == O_j \rrbracket \nonumber \\
	       & +  \mathbf{w}_2 \llbracket O_j == O_i + 1 \rrbracket +
	          \mathbf{w}_3 \llbracket O_j == O_i + 2 \rrbracket \nonumber \\
	       & +  \mathbf{w}_4 \llbracket O_j == O_i - 1 \rrbracket +
	          \mathbf{w}_5 \llbracket O_j == O_i - 2 \rrbracket \nonumber \\
	       & +  \mathbf{w}_6 \llbracket O_j \geq O_i + 3\ \textbf{ or }\ O_j \leq O_i - 3 \rrbracket 
	\label{penalty_obj}
\end{align}

%\marginpar{merge +3/-3 like in code, or rerun code} 
\noindent where $\llbracket \cdot \rrbracket$ is the Iverson bracket, it returns $1$ if the logical statement in the bracket is true, and $0$ otherwise. Note how for every pair of stops $i$ and $j$, only one of the Iverson brackets will return the value $1$, the others will be $0$. Hence, this function allows to put different penalties on different kinds of zone order violations.

%\subsubsection{Setting manual penalty parameters}

One approach to set penalty weights $w$ is to choose them manually. For example, if we want to produce routing plans prone to follow the sequencing while minimizing total travel time, we could set $\mathbf{w}_0 = 1, \mathbf{w}_1 = 0, \mathbf{w}_2 = 0$ and all other weights to $+\infty$ (or a very large value). 

Alternatively, the zone order penalties could decrease step-wise for larger gaps between the zones of subsequent stops. This could allow ``jumps'' between adjacent zones, as observed in a number of the actual expert solutions (see also data analysis in the experiment section). However, determining the right penalty values is tedious, prone to error and bias, and is even more difficult to keep up-to-date over time.

\subsubsection{Learning penalty parameters via SOP}

Given the weighted penalty function, just like in the zone-order TSP case, we can learn these weight penalties automatically on a training set. For this, we can again use Structured Output Prediction; more specifically, the structured perceptron algorithm described in the previous section.

We now provide a mapping to the required input of the perceptron algorithm:

\begin{itemize}
	\item $x$ is again a TSP solution, represented by an $n \times n$ adjancency matrix $x=[x_{ij}]$ (in this case, over the stops instead of over the zones);
	\item $u$ is the input structure, for which in this case we propose the tuple $u = (n, D, O)$ where $n$ is the number of stops in the instance, $D$ the travel time matrix between the $n$ stops, and $O$ the zone order index of the $n$ stops;
	\item $\mathbf{w}$ is the length-7 weight vector $\mathbf{w} = [w_0, \ldots, w_6]$ that balances the importance of the distance and the different penalties;
	\item $\Phi(u,x)$ is the length-7 feature vector which computes the total distance as well as the total number of violations of every type; we define
	\begin{align}
		\Phi\left(u,x\right) 
		&= \Phi\left((n, [d_{ij}], [O_{i}]), x\right) \\
		&= \left[\sum_{ij} T_{ij}x_{ij}, \sum_{ij} \llbracket O_i == O_j \rrbracket, \sum_{ij} \llbracket O_j == O_i + 1 \rrbracket, \ldots, \sum_{ij}  \llbracket O_j \geq O_i + 3\ \textbf{ or }\ O_j \leq O_i - 3 \rrbracket \right] \nonumber
	\end{align}
\end{itemize}

From this, as in the previous section, we can similarly derive
\begin{align*}
	\hat{x} &= \argmax_{x\in\mathcal{X}(u)} \mathbf{w}^\intercal\Phi(u,x) \\
	%&= \argmax_{y\in\mathcal{Y}(x)} w_d\sum_{ij} log(D'_{ij})y_{ij} + w_p\sum_{ij} log(\mathbf{P}_{ij})y_{ij} \\
	%&= \argmax_{y\in\mathcal{Y}(x)} \sum_{ij} (w_d log(D'_{ij}) + w_p log(\mathbf{P}_{ij}))y_{ij} \\
	%&= \argmin_{y\in\mathcal{Y}(x)} \sum_{ij} (-w_d log(D'_{ij}) - w_p log(\mathbf{P}_{ij}))y_{ij} \\
	&= \argmax_{x\in\mathcal{X}(u)} \sum_{ij} c_{ij} x_{ij}, \quad \mbox{with } c_{ij} ~\mbox{as defined in Equation~\eqref{penalty_obj}} \\
	&= TSP([c_{ij}]).
\end{align*}

We want to stress that although Equation~\eqref{penalty_obj} contains logical expressions, every $O_i$ is a constant for a given zone ordering $ZO,$ hence every computed $c_{ij}$ is also a constant. While the equation looks like a multi-objective formulation, and it is a multi-objective formulation, each of the sub-objectives is over the same $x_{ij}$ variables and hence this `multi' objective formulation can be equivalently rewritten as a single linear function over the $x_{ij}$ variables.

\section{Numerical Results}
%\textbf{T: do these answer the research quesiotns? Perhaps order your exps according to the RQs?}

%\begin{itemize}
%    \item Data description: Include all the statistics that are relevant to describe data: Division of High Medium Low; Zone ordering is hard or soft constraint, in the second case how many times zone ordering is violated; 
%    \item Training and testing set split
%    \item Explain evaluation metrics
%\end{itemize}

We now turn to evaluating the performance and benefits of our proposed approach.

\begin{figure}
    \centering
    \subfigure[Count of routes per stations]{
    \includegraphics[width=0.45\linewidth]{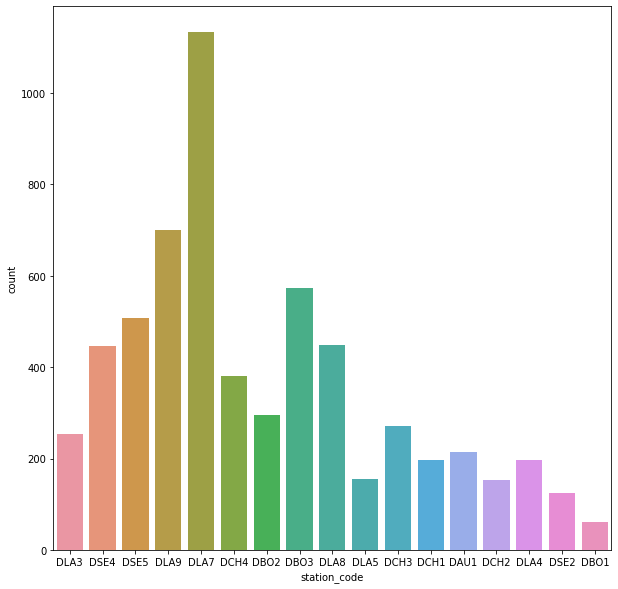}
    \label{fig:count_routes}
    }
    \subfigure[Stops overlap between stations]{
    \includegraphics[width=0.45\linewidth]{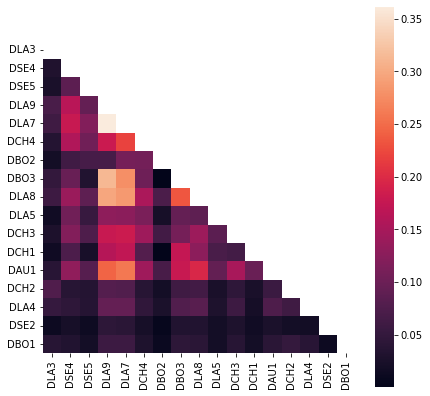}
    \label{fig:stops-overlap}
    }
    
\end{figure}

\paragraph{\bf Data description} For an empirical evaluation of the proposed methodology, we conduct our experiments on data provided by Amazon for the Last Mile Routing Challenge \footnote{\url{https://routingchallenge.mit.edu/}}. The provided data consists of 6112 historical routes. Each route is a feasible TSP solution which starts and ends at a station, and visits from 31 to 238 stops. As displayed on Fig.~\ref{fig:count_routes} the distribution of routes between the 17 stations is not uniform.  Fig.~\ref{fig:stops-overlap} shows how stops are shared between stations. All stops are clustered into 8868 zones. Planning experts assigned a quality score label (\texttt{high, medium, low}) to each route.%, assessing the amount of backtracking in the observed sequence. 
Vehicle capacities, times of departure and time windows of stops were not considered, although they can be added to the TSP solver without changes to the methodology. 

We can extract a \emph{zone ordering} from each route and empirically observe transitions between (non)adjacent zones.  Table~\ref{tab:zone_order_eda} denotes the frequency of such transitions. \texttt{next} and \texttt{prev} respectively stands for moving ahead and backward according to the zone ordering. \texttt{next2} and \texttt{prev2} denote moving two steps in the related direction. \texttt{next3+} and \texttt{prev3+} account for three or more steps, which is a more blatant zone order violation. 
As displayed on Table~\ref{tab:zone_order_eda}, those are more common in lower quality route. Medium-quality routes respectively have on average $0.073\%$ and $0.178\%$ of those transitions, against only $0.071\%$ and $0.168\%$ for high-quality routes. The small amount of low-quality routes available does not allow for a fair comparison. Nevertheless, this observation substantiates the underlying hypothesis that more zone order violations in a route lead to worse route quality, and that different penalties for different order violations are sensible. 

\paragraph{\bf Scoring} The Amazon challenge scoring function computes the similarity between the actual (zone or stop) sequence and an algorithm-produced sequence. The computation combines Sequence Deviation and Edit Distance with Real Penalty: 
\[
\mathtt{score} = \frac{SD(A, B) \cdot ERP_{norm}(A, B)}{ERP_e(A, B)},
\]
where $A$ is the actual sequence and $B$ is the user-submitted sequence. $SD$ denotes the Sequence Deviation of $B$ with respect to $A$, $ERP_{norm}$ denotes the Edit Distance with Real Penalty applied to sequences $A$ and $B$ with normalized distance or travel times, and $ERP_e$ denotes the number of edits prescribed by the  algorithm on sequence $B$ with respect to $A.$

\paragraph{\bf System description}
All experiments were run on a Lenovo ThinkPad X1 Carbon with an Intel Core i7 processor running at 1.8GHz with 16GB RAM.
As TSP solver, we used OR-Tools 9.0 with the guided local search as the search strategy.
During SOP training and to compute the final zone and stop TSPs that are scored, we set a timeout of 30 seconds to each TSP solver call. The timeout is reached for each TSP computation.
Probability estimation and the structured perceptron algorithm were implemented in \texttt{numpy}. We used a fixed learning rate of $\delta=10^{-5},$ determined after small-scale experiments on the training data. We train for $e=1$ epoch unless mentioned otherwise. \footnote{We will make our implementation available under an open source license upon acceptance of the paper.}

%\begin{figure}
%    \centering
%    \includegraphics[width=0.5\linewidth]{heatmap-overlap-station.png}
%    \caption{Stops overlap between stations}
%    \label{fig:stops-overlap}
%\end{figure}

\begin{table}[t]
    \centering
    \caption{Percentage of stops per route for each kind of zone order transition/violation, as described in Eq.~\ref{penalty_obj} }
    \vspace{1em}
    \begin{scriptsize}

    \begin{tabular}{lrrrrrrr}
    \toprule
    {} &  same &  next &  next2 &  next3+ &  prev &  prev2 &  prev3+   \\
    route quality label &          &            &            &             &                       &             &                      \\
    \midrule
    \texttt{high}        &  80.355\% &    13.552\% &      0.115\% &                 0.071\% &     0.209\% &      0.050\% &                0.168\% \\
    \texttt{medium}      &  89.295\% &    13.642\% &      0.107\% &                 0.073\% &     0.190\% &      0.048\% &                0.178\% \\
    \texttt{low}         &  87.610\% &    13.668\% &      0.120\% &                 0.053\% &     0.174\% &      0.033\% &                0.140\% \\
    \bottomrule
    \end{tabular}
    \end{scriptsize}
    \label{tab:zone_order_eda}
\end{table}
\paragraph{\bf Training and testing set split}
Note that we do not have access to the hidden dataset used in the challenge evaluation. We hence conduct our experiments on the historical data which we split into a train and a test set. As each of the 6112 historical instances is tagged with a route quality label, we employ \emph{stratified sampling} in order to sample each route label subpopulation (\texttt{high, medium, low}) independently. %In the evaluation (on the test set) we only use high-quality routes.

\begin{table}[htpb]
\centering
\begin{scriptsize}
\caption{Data partitioned into training and test sets \label{tab:data}}
\vspace{1em}

\begin{tabular}{@{}lr|rr@{}}
%\toprule
Route score & Total & Train & Test  \\ \midrule
\texttt{high}        & 2718 & 2174  & 544   \\
\texttt{medium}    & 3292  & 2634  & 658    \\
\texttt{low}       & 102  & 82    & 20     \\ \midrule
Total      & 6112 & 4890  & 1222   \\ %\bottomrule
\end{tabular}
\end{scriptsize}
\end{table}

% \begin{itemize}
%     \item Stage 1 experiments
%         \begin{itemize}
%             \item Comparison of zone sequencing accuracy (distance-based vs Markov vs SOP, etc.); default parameters
%             \item Numerical route score values tuning
%             \item SOP parameters tuning
%         \end{itemize}
% \end{itemize}

\begin{figure}[htpb]
    \centering
    \includegraphics[width=0.55\linewidth]{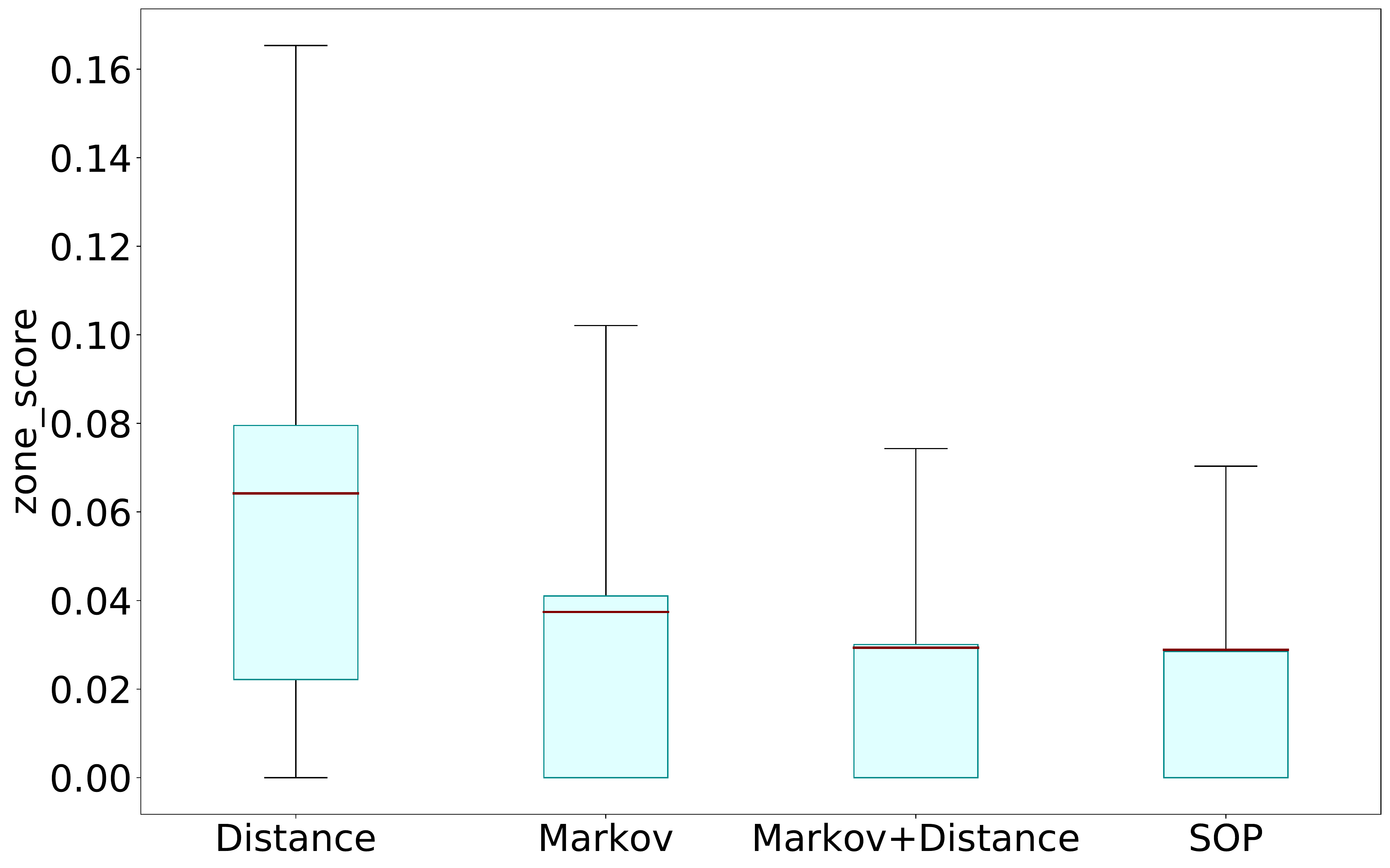}
    
    $ $
    
    \begin{scriptsize}
    \begin{tabular}{@{}ccccccc@{}}
    \toprule
    \multicolumn{3}{c}{Routes used for training} & \multicolumn{4}{c}{zone\_score} \\ \cmidrule(r){1-3} \cmidrule(l){4-7}
    \multicolumn{3}{c}{ } & Distance & Markov & Markov+Distance & SOP \\ \midrule
    \multicolumn{3}{c}{\texttt{high}} & 0.0642 & 0.0422 & 0.0314 & 0.0317 \\ 
    \multicolumn{3}{c}{\texttt{high} + \texttt{medium}} & 0.0642 & 0.0371 & 0.0293 & 0.0291 \\ 
    \multicolumn{3}{c}{\texttt{high} + \texttt{medium} + \texttt{low}} & 0.0642 & 0.0374 & 0.0291 & \textbf{0.0289} \\
    \bottomrule
    \end{tabular}
    
    \end{scriptsize}

    \caption{Zone scores of zone sequencing approaches on test data}
    \label{fig:zone_seq_comp}
\end{figure}

\begin{figure}[htpb]
    \centering
    \begin{scriptsize}
    
    \begin{tabular}{@{}cccc@{}}
    \toprule
    \# of epochs & $w_d$ & $w_p$ & zone\_score \\ \midrule
    0 & 1 & 1 & 0.0291 \\
    1 & 1.58 & 2.33 & 0.0289 \\
    2 & 2.16 & 3.66 & 0.0297 \\
    3 & 2.75 & 4.99 & 0.0296 \\
    4 & 3.33 & 6.32 & 0.0295 \\
    5 & 3.91 & 7.66 & 0.0295 \\ \bottomrule
    \end{tabular}
    
    \end{scriptsize}

    \caption{Trained weights and corresponding zone scores for different epochs of SOP}
    \label{fig:epochs}
\end{figure}

\paragraph{\bf Stage 1 experiments: zone-level ordering quality}

In the first set of experiments, we test and compare the performance of the different Stage 1 zone sequencing approaches that we presented in the previous section, namely, the distance-based (Distance), transition probabilities/preference-based (Markov), and the mixed distance and transition probabilities (SOP) approaches, where the component weights in the mixed approach are determined by structured prediction.

\begin{figure}[htpb]
\centering
\begin{minipage}{.5\textwidth}
  \centering
  \includegraphics[width=.97\linewidth]{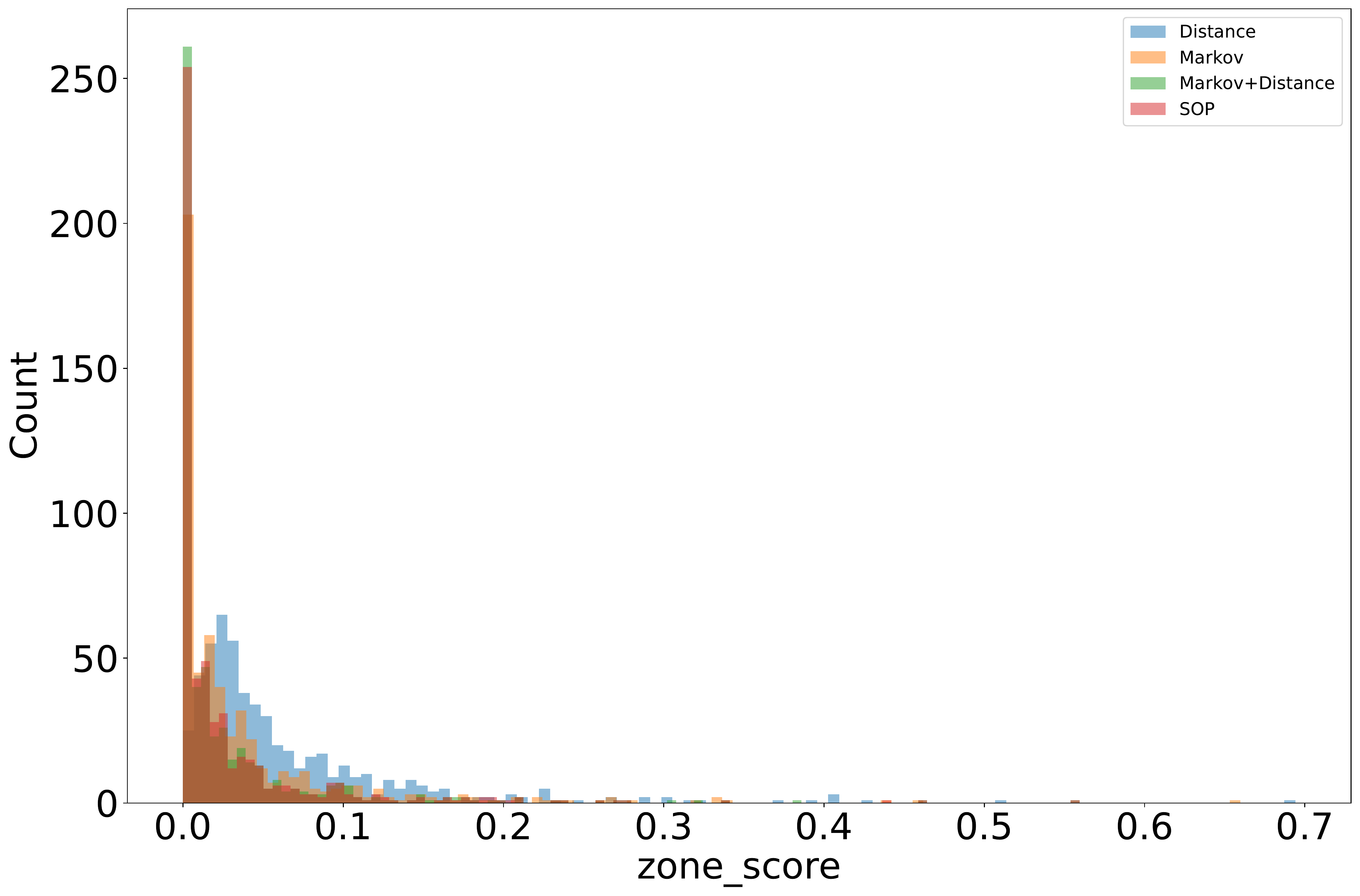}
  \label{fig:test1}
\end{minipage}%
\begin{minipage}{.5\textwidth}
  \centering
  \includegraphics[width=.95\linewidth]{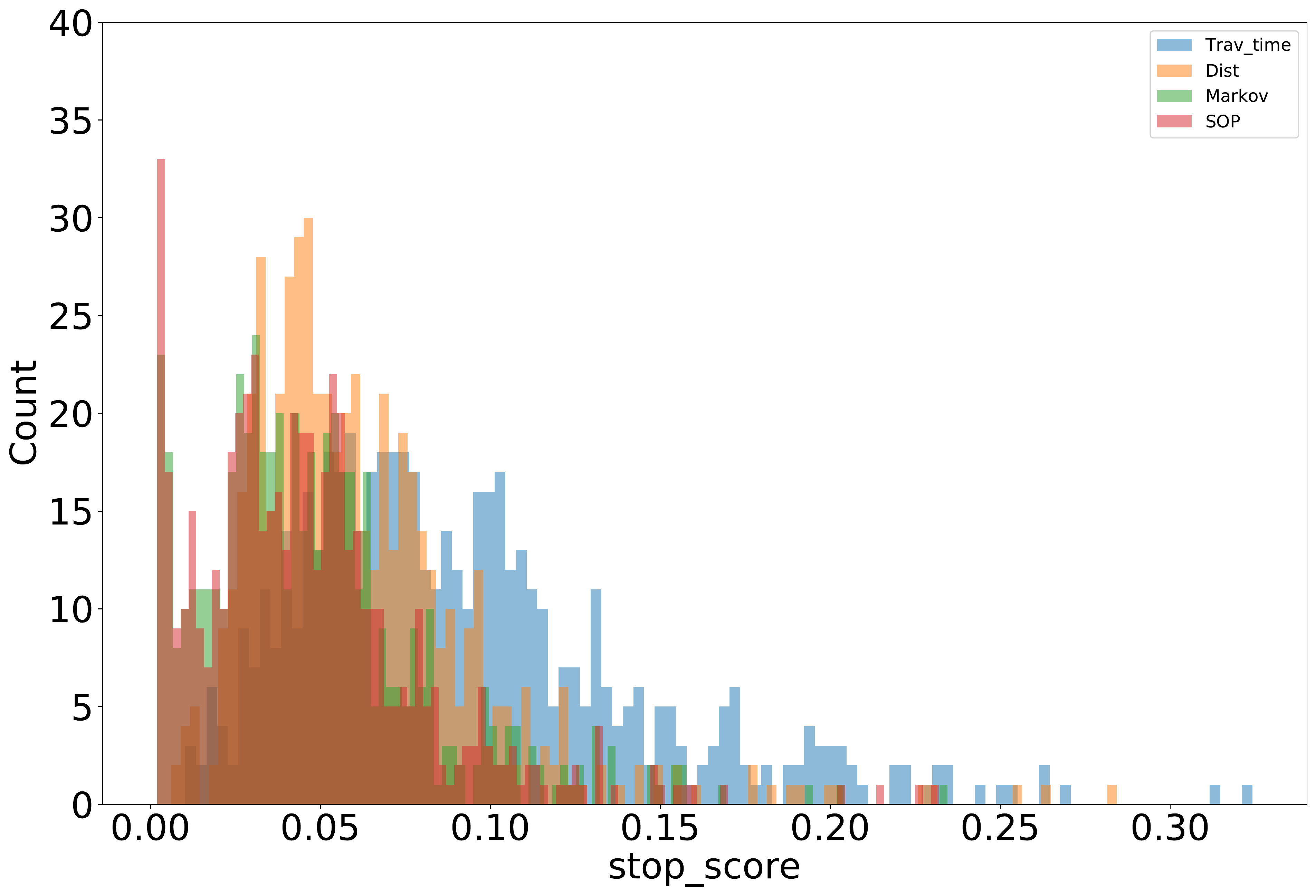}
  \label{fig:test2}
\end{minipage}
\caption{Histograms showing distribution of scores of the stop and zone sequencing approaches}
\label{fig:histograms}
\end{figure}

A graphical visualization of the distribution of the zone scores, as well as a summary of the average zone scores, for the four approaches are shown in Figure~\ref{fig:zone_seq_comp}. Visually, we can already notice a significant improvement when using the Markov model instead of the absolute distance-based approach. Furthermore, when distance and preference components are mixed with equal weight, we get further improved zone scores. The use of Structured Output Prediction (SOP) to find better weights marginally improves the result further. A histogram showing the frequency distribution of the zone scores of the four approaches is shown in Figure \ref{fig:histograms}.

In the table at the bottom of the figure we show the average zone score when using only the \texttt{high} quality labeled routes, or also the \texttt{medium} and \texttt{low}. Although they differ in quality, we can see that using all data results in better overall zone orders. We believe this is due simply to the \textit{amount} of data available, as can be seen from Table~\ref{tab:data}.

We take a closer look at the performance of the SOP approach in Figure~\ref{fig:epochs}. For this experiment, we ran the structured perceptron algorithm for multiple epochs. In the figure, we can see that starting from uniform weights, the zone score on the test data slightly increases before it worsens again a bit. Looking at the (interpretable) weights, we see that learning especially increases the weight of the prefence matrix compared to the weight of the distance matrix. Runtimes are not shown, but we observed that every instance reaches the timeout of 30 seconds during SOP training.

\begin{figure}[htpb]
    \centering
    \includegraphics[width=0.60\linewidth]{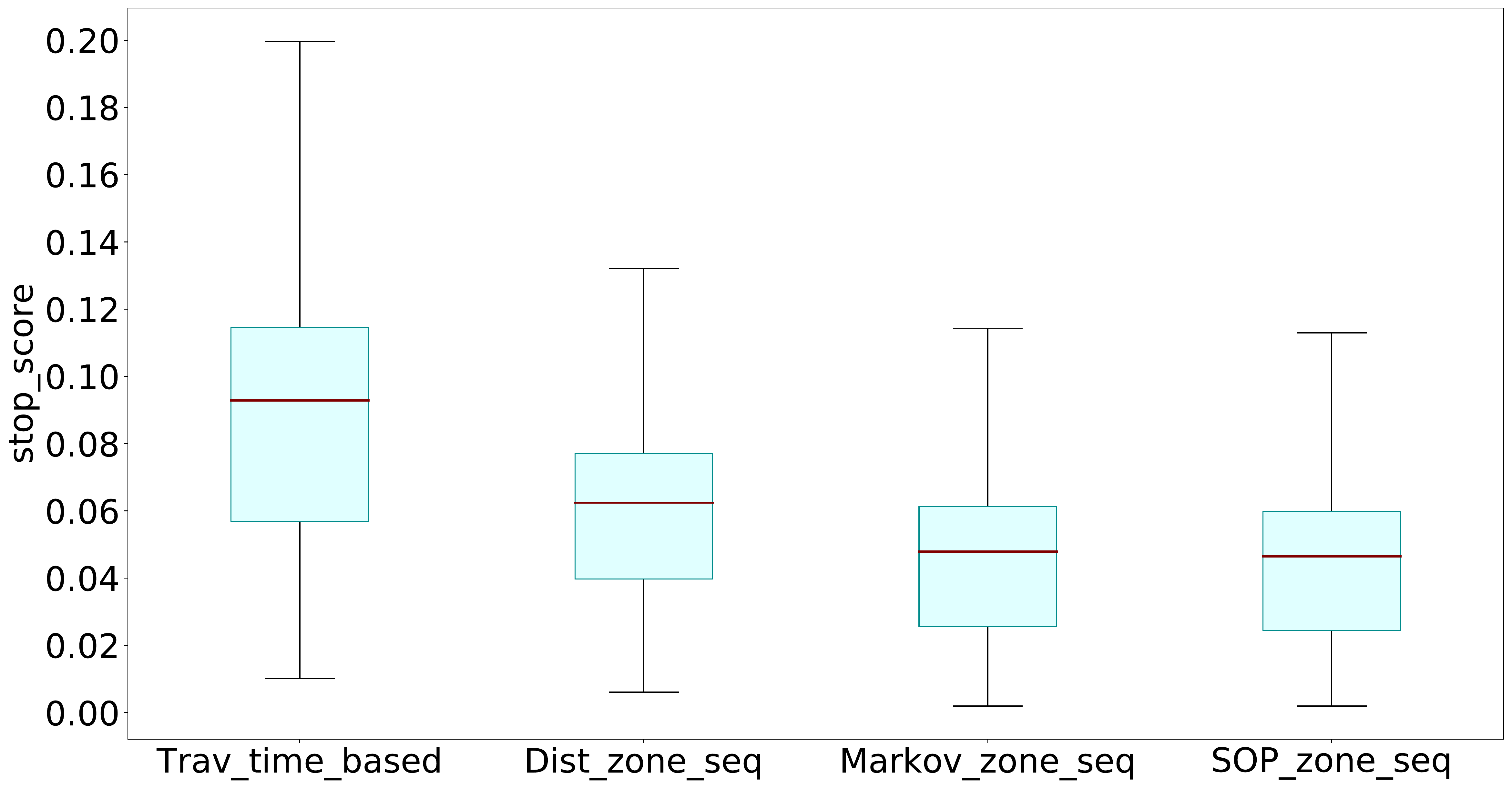}

    $ $
    
    \begin{scriptsize}

    % \begin{tabular}{@{}lrrrr@{}}
    % \toprule
    %             & Trav\_time\_based & Dist-based zone seq & Markov-based zone seq & SOP-based zone seq \\ \cmidrule(l){3-5} 
    %             &           & \multicolumn{3}{c}{(zone order as hard constraint)} \\ \midrule
    % zone\_score & ---       & 0.0644     & 0.0374   & 0.0291    \\ 
    % stp\_sc and zone\_sc correlation & ---       & 0.5424     & 0.5273   & 0.5454    \\
    % stop\_score & 0.0929    & 0.0618     & 0.0511   & 0.0504      \\ \midrule 
    %  &           & \multicolumn{3}{c}{(slightly relaxed zone order constraint)} \\ \cmidrule(l){3-5}
    %  stop\_score &     & 0.0618     & 0.0510   & 0.0503      \\  \midrule
    %  &           & \multicolumn{3}{c}{(with parameters by SOP)} \\ \cmidrule(l){3-5}
    %  stop\_score &     & 0.0630     & 0.0484   & 0.0479      \\
    % \bottomrule
    % \end{tabular}

    % \begin{tabular}{@{}lcccc@{}}
    % \toprule
    %  & Trav\_time-based & Dist-based zone seq & Markov-based zone seq & SOP-based zone seq \\ \midrule
    % Zone order as hard constraint & 0.0929 & 0.0618 & 0.0511 & 0.0504 \\
    % Relaxed zone order constraint & 0.0929 & 0.0618 & 0.0510 & 0.0503 \\
    % Penalty parameters by SOP & 0.0929 & 0.0630 & 0.0484 & 0.0479 \\ \bottomrule
    % \end{tabular}

    \begin{tabular}{@{}ccccccccccc@{}}
    \toprule
    \multicolumn{7}{c}{Penalty weights} &  & \multicolumn{3}{c}{Zone sequencing approach} \\ \cmidrule(){1-7} \cmidrule(lr){8-8} \cmidrule(){9-11}
    $w_0$ & $w_1$ & $w_2$ & $w_3$ & $w_4$ & $w_5$ & $w_6$ & Trav\_time-based & Dist & Markov & SOP \\ \midrule
     2 & 1 & 1.5 & 2 & 1.5 & 2 & 2 & 0.0929 & 0.0617 & 0.0510 & 0.0501 \\
     2 & 1 & 2 & 4 & 2 & 4 & 6 & 0.0929 & 0.0618 & 0.0507 & 0.0498 \\
     2 & 0.35 & 2.13 & 4.09 & 2.32 & 4.02 & 6.10 & 0.0929 & 0.0625 & 0.0479 & \textbf{0.0465} \\ \bottomrule
    \end{tabular}

    \end{scriptsize}

    \caption{Stop scores of stop sequencing approaches on test data}
    \label{fig:stops_seq_comp_def}
\end{figure}

% \begin{figure}[htpb]
%     \centering
%     \includegraphics[width=0.8\linewidth]{Stage2Histogram_hard.pdf}
% \end{figure}

\paragraph{\bf Stage 2: stop-level experiments}

The different stage 2 stop ordering approaches are evaluated and compared in this next set of experiments. The first (\texttt{Trav\_time\_based}) is the conventional approach of solving (TSP) with the given travel time matrix as the cost matrix. The other three are \emph{zone penalty} approaches, where the zone penalty is determined by the zone order generated in Stage 1.

Numerical results are summarized in Figure~\ref{fig:stops_seq_comp_def}. 
The figure summarizes the results, showing that a non-zone based approach using distances performs worst. Even when only using distances, but in a two-stage approach, e.g., zone penalties from a distance-based zone ordering, already leads to improved results. This is in line with our data analysis that a zone ordering is typically followed. Our next two approaches, Markov\_zone\_seq and SOP\_zone\_seq differ only in how the zone ordering was computed. As their zone orderings performed rather similar, we also see here that their stop-level scores are also rather similar.

We now look at the effect of different zone penalty weights, at the bottom of Fig~\ref{fig:stops_seq_comp_def}. We tried two hand-crafted penalty values: one in which the penalty for `larger' zone violations gradually increases (1, 1.5, 2) and one in which it more rapidly increases (1,2,4,6). Finally, the bottom line, which was initialized with the better-performing second row values, shows the output after 1 epoch of SOP learning on these weights. We can see that the best approach is the combination of Markov estimation with SOP both at the zone-level and at the stop-level. Looking at the bottom entry in particular, we can see that the stop-level SOP especially learned to decrease the same-zone penalty weight, as well as increasing `previous' and `next' penalty scores. The histogram in Figure \ref{fig:histograms} shows the frequency distribution of the stop scores of the four approaches, where most of the scores of the better preforming methods, SOP and Markov, tend to be closer to zero.
%For the \emph{zone penalty} approaches, \texttt{Dist\_zone\_seq} represents the stop sequencing approach where the zone order is determined by distance-based zone sequencing in Stage 1, etc. 

\section{Concluding Remarks}

This article studies the problem of learning the preferences of drivers and planners over a set of delivery locations in the last mile delivery context. The main difficulty of this problem is that clients do not appear often in the historical data, making it impossible to learn transition preferences at the stop level. Hence, we study a setting where stops are clustered into zones. We propose a two-stage TSP approach to solve this problem. 

In the first stage, the preferences of transitions between zones are learned from historical data. This allows us to produce sequences of zones aligned to the preferences of the expert users. We adapt from literature a Markovian approach to learn zone transition probabilities and combine these with pairwise zone distances to generate zone sequences that not only minimize the travel distance but also maximizes user preferences. Furthermore, we apply the structured output prediction approach to learn the optimal parameters to be used in combining the distances and the transition probabilities.

In the second stage, we structure the stop routing problem into a TSP whose objective function combines the travel time and zone transition penalties derived from the zone order obtained in the previous stage. As the objective function now takes the form of a weighted multi-objective function, we again see the opportunity to apply structured prediction to fine-tune some initial penalty values.

Our computational results show that compared to the conventional method of stop routing by TSP using travel times, our two-stage zoning approach significantly performs better. Stage 1 experiments confirm that combining distances and transition probabilities result to better zone sequences. Stage 2 experiments show that our proposed zone penalty framework results to higher quality stop sequences. In both sets of experiments, we confirm that we can obtain improved results by using structured output prediction. 

Future work will involve applications to other extended settings, such as vehicle routing. Another relevant future research question is to measure the impact in accuracy and computational times of using different methods to solve the TSP either exactly or by heuristics; as well as \textit{incremental} methods that do not require to re-solve from scratch in every structured-output prediction loop. Another avenue that can be explored is additional features to be used for learning, both at the zone level and the stop level, as well as the use of other structured output prediction methods including deep learning ones. However, more complex learning architectures may also require more training data.

%The potential of getting better results by using other TSP meta heuristics is also worth investigating. 
%%% EXCLUDED FROM THE PAGE LIMIT: %%%%
% \clearpage
% \appendix
% \section{Optional Appendix}
% If needed, you can provide supplementary information in the form of an appendix, which does not count against the page limit. 
% However, do not include anything in the appendix that is critical for the understanding of your model and its performance.

\bibliography{mybibfile}

\begin{thebibliography}{22}
\expandafter\ifx\csname natexlab\endcsname\relax\def\natexlab#1{#1}\fi
\providecommand{\url}[1]{\texttt{#1}}
\providecommand{\href}[2]{#2}
\providecommand{\path}[1]{#1}
\providecommand{\DOIprefix}{doi:}
\providecommand{\ArXivprefix}{arXiv:}
\providecommand{\URLprefix}{URL: }
\providecommand{\Pubmedprefix}{pmid:}
\providecommand{\doi}[1]{\href{http://dx.doi.org/#1}{\path{#1}}}
\providecommand{\Pubmed}[1]{\href{pmid:#1}{\path{#1}}}
\providecommand{\bibinfo}[2]{#2}
\ifx\xfnm\relax \def\xfnm[#1]{\unskip,\space#1}\fi
%Type = Book
\bibitem[{Applegate et~al.(2011)Applegate, Bixby, Chv{\'a}tal \&
  Cook}]{applegate2011traveling}
\bibinfo{author}{Applegate, D.~L.}, \bibinfo{author}{Bixby, R.~E.},
  \bibinfo{author}{Chv{\'a}tal, V.}, \& \bibinfo{author}{Cook, W.~J.}
  (\bibinfo{year}{2011}).
\newblock {\it \bibinfo{title}{The traveling salesman problem}\/}.
\newblock \bibinfo{publisher}{Princeton university press}.
%Type = Book
\bibitem[{BakIr et~al.(2007)BakIr, Hofmann, Schölkopf, Smola, Taskar \&
  Vishwanathan}]{sop_book}
\bibinfo{author}{BakIr, G.}, \bibinfo{author}{Hofmann, T.},
  \bibinfo{author}{Schölkopf, B.}, \bibinfo{author}{Smola, A.~J.},
  \bibinfo{author}{Taskar, B.}, \& \bibinfo{author}{Vishwanathan, S.}
  (\bibinfo{year}{2007}).
\newblock {\it \bibinfo{title}{{Predicting Structured Data}}\/}.
\newblock \bibinfo{publisher}{The MIT Press}.
\newblock \URLprefix \url{https://doi.org/10.7551/mitpress/7443.001.0001}.
  \DOIprefix\doi{10.7551/mitpress/7443.001.0001}.
%Type = Inproceedings
\bibitem[{Canoy \& Guns(2019)}]{canoy2019CP}
\bibinfo{author}{Canoy, R.}, \& \bibinfo{author}{Guns, T.}
  (\bibinfo{year}{2019}).
\newblock \bibinfo{title}{Vehicle routing by learning from historical
  solutions}.
\newblock In {\it \bibinfo{booktitle}{International Conference on Principles
  and Practice of Constraint Programming}\/} (pp. \bibinfo{pages}{54--70}).
\newblock \bibinfo{organization}{Springer}.
%Type = Inproceedings
\bibitem[{Ceikute \& Jensen(2013)}]{ceikute2013routing}
\bibinfo{author}{Ceikute, V.}, \& \bibinfo{author}{Jensen, C.~S.}
  (\bibinfo{year}{2013}).
\newblock \bibinfo{title}{Routing service quality--local driver behavior versus
  routing services}.
\newblock In {\it \bibinfo{booktitle}{2013 IEEE 14th International Conference
  on Mobile Data Management}\/} (pp. \bibinfo{pages}{97--106}).
\newblock \bibinfo{organization}{IEEE} volume~\bibinfo{volume}{1}.
%Type = Article
\bibitem[{Chen et~al.(2021)Chen, Chen \& Langevin}]{chen2021inverse}
\bibinfo{author}{Chen, L.}, \bibinfo{author}{Chen, Y.}, \&
  \bibinfo{author}{Langevin, A.} (\bibinfo{year}{2021}).
\newblock \bibinfo{title}{An inverse optimization approach for a capacitated
  vehicle routing problem}.
\newblock {\it \bibinfo{journal}{European Journal of Operational Research}\/},
  .
%Type = Inproceedings
\bibitem[{Chen et~al.(2020)Chen, Li, Umarov, Gao \& Song}]{Chen0UGS20}
\bibinfo{author}{Chen, X.}, \bibinfo{author}{Li, Y.}, \bibinfo{author}{Umarov,
  R.}, \bibinfo{author}{Gao, X.}, \& \bibinfo{author}{Song, L.}
  (\bibinfo{year}{2020}).
\newblock \bibinfo{title}{{RNA} secondary structure prediction by learning
  unrolled algorithms}.
\newblock In {\it \bibinfo{booktitle}{8th International Conference on Learning
  Representations, {ICLR} 2020, Addis Ababa, Ethiopia, April 26-30, 2020}\/}.
%Type = Inproceedings
\bibitem[{Clarke et~al.(2010)Clarke, Goldwasser, Chang \& Roth}]{ClarkeGCR10}
\bibinfo{author}{Clarke, J.}, \bibinfo{author}{Goldwasser, D.},
  \bibinfo{author}{Chang, M.}, \& \bibinfo{author}{Roth, D.}
  (\bibinfo{year}{2010}).
\newblock \bibinfo{title}{Driving semantic parsing from the world's response}.
\newblock In \bibinfo{editor}{M.~Lapata}, \& \bibinfo{editor}{A.~Sarkar}
  (Eds.), {\it \bibinfo{booktitle}{Proceedings of the Fourteenth Conference on
  Computational Natural Language Learning, CoNLL 2010, Uppsala, Sweden, July
  15-16, 2010}\/} (pp. \bibinfo{pages}{18--27}).
\newblock \bibinfo{publisher}{{ACL}}.
%Type = Inproceedings
\bibitem[{Collins(2002)}]{collins2002discriminative}
\bibinfo{author}{Collins, M.} (\bibinfo{year}{2002}).
\newblock \bibinfo{title}{Discriminative training methods for hidden markov
  models: Theory and experiments with perceptron algorithms}.
\newblock In {\it \bibinfo{booktitle}{Proceedings of the 2002 conference on
  empirical methods in natural language processing (EMNLP 2002)}\/} (pp.
  \bibinfo{pages}{1--8}).
%Type = Article
\bibitem[{Dantzig \& Ramser(1959)}]{dantzig1959truck}
\bibinfo{author}{Dantzig, G.~B.}, \& \bibinfo{author}{Ramser, J.~H.}
  (\bibinfo{year}{1959}).
\newblock \bibinfo{title}{The truck dispatching problem}.
\newblock {\it \bibinfo{journal}{Management science}\/},  {\it
  \bibinfo{volume}{6}\/}, \bibinfo{pages}{80--91}.
%Type = Inproceedings
\bibitem[{Delling et~al.(2015)Delling, Goldberg, Goldszmidt, Krumm, Talwar \&
  Werneck}]{delling2015navigation}
\bibinfo{author}{Delling, D.}, \bibinfo{author}{Goldberg, A.~V.},
  \bibinfo{author}{Goldszmidt, M.}, \bibinfo{author}{Krumm, J.},
  \bibinfo{author}{Talwar, K.}, \& \bibinfo{author}{Werneck, R.~F.}
  (\bibinfo{year}{2015}).
\newblock \bibinfo{title}{Navigation made personal: Inferring driving
  preferences from gps traces}.
\newblock In {\it \bibinfo{booktitle}{Proceedings of the 23rd SIGSPATIAL
  international conference on advances in geographic information systems}\/}
  (pp. \bibinfo{pages}{1--9}).
%Type = Inproceedings
\bibitem[{Joachims(2002)}]{joachims2002optimizing}
\bibinfo{author}{Joachims, T.} (\bibinfo{year}{2002}).
\newblock \bibinfo{title}{Optimizing search engines using clickthrough data}.
\newblock In {\it \bibinfo{booktitle}{Proceedings of the eighth ACM SIGKDD
  international conference on Knowledge discovery and data mining}\/} (pp.
  \bibinfo{pages}{133--142}).
%Type = Inproceedings
\bibitem[{Kool et~al.(2019)Kool, van Hoof \& Welling}]{kool2018attention}
\bibinfo{author}{Kool, W.}, \bibinfo{author}{van Hoof, H.}, \&
  \bibinfo{author}{Welling, M.} (\bibinfo{year}{2019}).
\newblock \bibinfo{title}{Attention, learn to solve routing problems!}
\newblock In {\it \bibinfo{booktitle}{International Conference on Learning
  Representations}\/}.
\newblock \URLprefix \url{https://openreview.net/forum?id=ByxBFsRqYm}.
%Type = Article
\bibitem[{Laporte(1992)}]{laporte1992traveling}
\bibinfo{author}{Laporte, G.} (\bibinfo{year}{1992}).
\newblock \bibinfo{title}{The traveling salesman problem: An overview of exact
  and approximate algorithms}.
\newblock {\it \bibinfo{journal}{European Journal of Operational Research}\/},
  {\it \bibinfo{volume}{59}\/}, \bibinfo{pages}{231--247}.
%Type = Inproceedings
\bibitem[{Letchner et~al.(2006)Letchner, Krumm \& Horvitz}]{letchner2006trip}
\bibinfo{author}{Letchner, J.}, \bibinfo{author}{Krumm, J.}, \&
  \bibinfo{author}{Horvitz, E.} (\bibinfo{year}{2006}).
\newblock \bibinfo{title}{Trip router with individualized preferences (trip):
  Incorporating personalization into route planning}.
\newblock In {\it \bibinfo{booktitle}{AAAI}\/} (pp.
  \bibinfo{pages}{1795--1800}).
%Type = Masterthesis
\bibitem[{Li \& Phillips(2018)}]{li2018learning}
\bibinfo{author}{Li, Y.}, \& \bibinfo{author}{Phillips, W.}
  (\bibinfo{year}{2018}).
\newblock {\it \bibinfo{title}{Learning from route plan deviation in last-mile
  delivery}\/}.
\newblock Master's thesis Massachusetts Institute of Technology, Cambridge.
%Type = Article
\bibitem[{Lu et~al.(2015)Lu, Wu, Mao, Wang \& Zhang}]{lu2015recommender}
\bibinfo{author}{Lu, J.}, \bibinfo{author}{Wu, D.}, \bibinfo{author}{Mao, M.},
  \bibinfo{author}{Wang, W.}, \& \bibinfo{author}{Zhang, G.}
  (\bibinfo{year}{2015}).
\newblock \bibinfo{title}{Recommender system application developments: a
  survey}.
\newblock {\it \bibinfo{journal}{Decision Support Systems}\/},  {\it
  \bibinfo{volume}{74}\/}, \bibinfo{pages}{12--32}.
%Type = Inproceedings
\bibitem[{Mandi et~al.(2021)Mandi, Canoy, Bucarey \& Guns}]{mandi2021data}
\bibinfo{author}{Mandi, J.}, \bibinfo{author}{Canoy, R.},
  \bibinfo{author}{Bucarey, V.}, \& \bibinfo{author}{Guns, T.}
  (\bibinfo{year}{2021}).
\newblock \bibinfo{title}{Data driven vrp: A neural network model to learn
  hidden preferences for vrp}.
\newblock In {\it \bibinfo{booktitle}{27th International Conference on
  Principles and Practice of Constraint Programming (CP 2021)}\/}
  (p.~\bibinfo{pages}{42}).
\newblock \bibinfo{organization}{Schloss Dagstuhl-Leibniz-Zentrum f{\"u}r
  Informatik}.
%Type = Article
\bibitem[{Miller et~al.(1960)Miller, Tucker \& Zemlin}]{miller1960integer}
\bibinfo{author}{Miller, C.~E.}, \bibinfo{author}{Tucker, A.~W.}, \&
  \bibinfo{author}{Zemlin, R.~A.} (\bibinfo{year}{1960}).
\newblock \bibinfo{title}{Integer programming formulation of traveling salesman
  problems}.
\newblock {\it \bibinfo{journal}{Journal of the ACM (JACM)}\/},  {\it
  \bibinfo{volume}{7}\/}, \bibinfo{pages}{326--329}.
%Type = Incollection
\bibitem[{Toledo et~al.(2013)Toledo, Sun, Rosa, Ben-Akiva, Flanagan, Sanchez \&
  Spissu}]{toledo2013decision}
\bibinfo{author}{Toledo, T.}, \bibinfo{author}{Sun, Y.}, \bibinfo{author}{Rosa,
  K.}, \bibinfo{author}{Ben-Akiva, M.}, \bibinfo{author}{Flanagan, K.},
  \bibinfo{author}{Sanchez, R.}, \& \bibinfo{author}{Spissu, E.}
  (\bibinfo{year}{2013}).
\newblock \bibinfo{title}{Decision-making process and factors affecting truck
  routing}.
\newblock In {\it \bibinfo{booktitle}{Freight Transport Modelling}\/}.
\newblock \bibinfo{publisher}{Emerald Group Publishing Limited}.
%Type = Inproceedings
\bibitem[{Vinyals et~al.(2015)Vinyals, Fortunato \& Jaitly}]{NIPS2015_29921001}
\bibinfo{author}{Vinyals, O.}, \bibinfo{author}{Fortunato, M.}, \&
  \bibinfo{author}{Jaitly, N.} (\bibinfo{year}{2015}).
\newblock \bibinfo{title}{Pointer networks}.
\newblock In \bibinfo{editor}{C.~Cortes}, \bibinfo{editor}{N.~Lawrence},
  \bibinfo{editor}{D.~Lee}, \bibinfo{editor}{M.~Sugiyama}, \&
  \bibinfo{editor}{R.~Garnett} (Eds.), {\it \bibinfo{booktitle}{Advances in
  Neural Information Processing Systems}\/}.
\newblock \bibinfo{publisher}{Curran Associates, Inc.}
  volume~\bibinfo{volume}{28}.
\newblock \URLprefix
  \url{https://proceedings.neurips.cc/paper/2015/file/29921001f2f04bd3baee84a12e98098f-Paper.pdf}.
%Type = Inproceedings
\bibitem[{Yadollahpour et~al.(2013)Yadollahpour, Batra \&
  Shakhnarovich}]{YadollahpourBS13}
\bibinfo{author}{Yadollahpour, P.}, \bibinfo{author}{Batra, D.}, \&
  \bibinfo{author}{Shakhnarovich, G.} (\bibinfo{year}{2013}).
\newblock \bibinfo{title}{Discriminative re-ranking of diverse segmentations}.
\newblock In {\it \bibinfo{booktitle}{2013 {IEEE} Conference on Computer Vision
  and Pattern Recognition, Portland, OR, USA, June 23-28, 2013}\/} (pp.
  \bibinfo{pages}{1923--1930}).
\newblock \bibinfo{publisher}{{IEEE} Computer Society}.
%Type = Article
\bibitem[{Zeng et~al.(2019)Zeng, Tong \& Chen}]{zeng2019last}
\bibinfo{author}{Zeng, Y.}, \bibinfo{author}{Tong, Y.}, \&
  \bibinfo{author}{Chen, L.} (\bibinfo{year}{2019}).
\newblock \bibinfo{title}{Last-mile delivery made practical: An efficient route
  planning framework with theoretical guarantees}.
\newblock {\it \bibinfo{journal}{Proceedings of the VLDB Endowment}\/},  {\it
  \bibinfo{volume}{13}\/}, \bibinfo{pages}{320--333}.

\end{thebibliography}
%\bibliographystyle{alpha} % says it is illegal to define one, maybe in template already

% Acknowledgments here
\section*{Acknowledgments}
This research received partial funding from the FWO Flanders project grant FWO-S007318N (Data-driven logistics), the European Research Council (ERC H2020, Grant agreement No. 101002802, CHAT-Opt), and the Institute for the encouragement of Scientific Research \& Innovation of Brussels (Innoviris, 2021-RECONCILE).
% Leave this (end of acknowledgment)

\end{document}